%% file: main.tex
\documentclass{article}

\usepackage{amsmath}
\usepackage{graphicx}
\usepackage{accents}
\usepackage{xcolor}
\usepackage{bbm}
\usepackage{float}
\usepackage{mdwlist}
\usepackage{multirow}
\usepackage{enumitem}
\usepackage{soul}
\usepackage{hyperref}
\usepackage{subfig}
\usepackage{tikz}
\usepackage{booktabs}
\usepackage{wrapfig}
\usepackage{xspace}
\usetikzlibrary{calc}

\makeatletter
\DeclareRobustCommand{\cev}[1]{%
  \mathpalette\do@cev{#1}%
}
\newcommand{\do@cev}[2]{%
  \fix@cev{#1}{+}%
  \reflectbox{$\m@th#1\vec{\reflectbox{$\fix@cev{#1}{-}\m@th#1#2\fix@cev{#1}{+}$}}$}%
  \fix@cev{#1}{-}%
}

\newcommand{\fix@cev}[2]{%
  \ifx#1\displaystyle
    \mkern#23mu
  \else
    \ifx#1\textstyle
      \mkern#23mu
    \else
      \ifx#1\scriptstyle
        \mkern#22mu
      \else
        \mkern#22mu
      \fi
    \fi
  \fi
}

\newcommand{\beginsupplement}{%
        \setcounter{table}{0}
        \renewcommand{\thetable}{S\arabic{table}}%
        \setcounter{figure}{0}
        \renewcommand{\thefigure}{S\arabic{figure}}%
     }
     
\newcommand\blfootnote[1]{%
  \begingroup
  \renewcommand\thefootnote{}\footnote{#1}%
  \addtocounter{footnote}{-1}%
  \endgroup
}

\makeatother

    \PassOptionsToPackage{numbers, compress, sort}{natbib}

\usepackage[preprint]{neurips_2019}

\usepackage[utf8]{inputenc} 
\usepackage[T1]{fontenc}    
\usepackage{hyperref}       
\usepackage{url}            
\usepackage{booktabs}       
\usepackage{amsfonts}       
\usepackage{nicefrac}       
\usepackage{microtype}      

\newcommand{\etal}{et al. }
\newcommand{\benchmarkName}{TAPE\xspace}

\title{Evaluating Protein Transfer Learning with TAPE}

\author{Roshan Rao$^{*1}$ \quad Nicholas Bhattacharya$^{*1}$ \quad 
Neil Thomas$^{*1}$ \quad\\
\textbf{Yan Duan}$^2$ \quad
\textbf{Xi Chen}$^2$ \quad
\textbf{John Canny}$^{1,3}$ \quad
\textbf{Pieter Abbeel}$^{1,2}$ \quad
\textbf{Yun S. Song}$^{1,4}$}

\begin{document}

\maketitle
\begin{abstract}
Protein modeling is an increasingly popular area of machine learning research. Semi-supervised learning has emerged as an important paradigm in protein modeling due to the high cost of acquiring supervised protein labels, but the current literature is fragmented when it comes to datasets and standardized evaluation techniques.
To facilitate progress in this field, we introduce the Tasks Assessing Protein Embeddings (\benchmarkName), a set of five biologically relevant semi-supervised learning tasks spread across different domains of protein biology. 
We curate tasks into specific training, validation, and test splits to ensure that each task tests biologically relevant generalization that transfers to real-life scenarios.
We benchmark a range of approaches to semi-supervised protein representation learning, which span recent work as well as canonical sequence learning techniques. 
We find that self-supervised pretraining is helpful for almost all models on all tasks, more than doubling performance in some cases. 
Despite this increase, in several cases features learned by self-supervised pretraining still lag behind features extracted by state-of-the-art  non-neural techniques. This gap in performance suggests a huge opportunity for innovative architecture design and improved modeling paradigms that better capture the signal in biological sequences. \benchmarkName will help the machine learning community focus effort on scientifically relevant problems. 
Toward this end, all data and code used to run these experiments are available at \url{https://github.com/songlab-cal/tape}.
\blfootnote{$^*$Equal Contribution $^1$UC Berkeley $^2$covariant.ai $^3$Google $^4$Chan Zuckerberg Biohub}
\blfootnote{Correspondence to: \texttt{\{roshan\_rao,nick\_bhat,nthomas,yss\}@berkeley.edu}}
\end{abstract}

\input{introduction.tex}
\input{background.tex}
\input{related_work.tex}
\input{datasets.tex}
\input{models.tex}

\input{results.tex}
\input{discussion.tex}
\input{future_work.tex}
\input{acknowledgements.tex}
\clearpage
\bibliography{main.bib}
\newpage
\input{supplement.tex}

\end{document}

%% file: introduction.tex
\section{Introduction}
\label{sec:introduction}

New sequencing technologies have led to an explosion in the size of protein databases over the past decades. These databases have seen exponential growth, with the total number of sequences doubling every two years \cite{uniprot}. Obtaining meaningful labels and annotations for these sequences requires significant investment of experimental resources, as well as scientific expertise, resulting in an exponentially growing gap between the size of protein sequence datasets and the size of annotated subsets. 
Billions of years of evolution have sampled the portions of protein sequence space that are relevant to life, so large unlabeled datasets of protein sequences are expected to contain significant biological information \cite{anfinsen, yanofsky, altschuh}. 
Advances in natural language processing (NLP) have shown that self-supervised learning is a powerful tool for extracting information from unlabeled sequences \cite{elmo, bert, gpt2}, which raises a tantalizing question: can we adapt NLP-based techniques to extract useful biological information from massive sequence datasets?

To help answer this question, we introduce the Tasks Assessing Protein Embeddings (\benchmarkName), which to our knowledge is the first attempt at systematically evaluating semi-supervised learning on protein sequences. \benchmarkName includes a set of five biologically relevant supervised tasks that evaluate the performance of learned protein embeddings across diverse aspects of protein understanding.

We choose our tasks to highlight three major areas of protein biology where self-supervision can facilitate scientific advances: structure prediction, detection of remote homologs, and protein engineering. We constructed data splits to simulate biologically relevant generalization, such as a model's ability to generalize to entirely unseen portions of sequence space, or to finely resolve small portions of sequence space. Improvement on these tasks range in application, including designing new antibodies \cite{antibodyengineering}, improving cancer diagnosis \cite{pancreatic-cancer-paper}, and finding new antimicrobial genes hiding in the so-called ``Dark Proteome'': tens of millions of sequences with no labels where existing techniques for determining protein similarity fail  \cite{darkproteome}.

We assess the performance of three representative models (recurrent, convolutional, and attention-based) that have performed well for sequence modeling in other fields to determine their potential for protein learning. We also compare two recently proposed semi-supervised models (Bepler \etal \cite{bepler}, Alley \etal \cite{unirep}). With our benchmarking framework, these models can be compared directly to one another for the first time.

We show that self-supervised pretraining improves performance for almost all models on all downstream tasks. Interestingly, performance for each architecture varies significantly across tasks, highlighting the need for a multi-task benchmark such as ours. We also show that non-deep alignment-based features \cite{psi-blast, hhblits, hhpred, profilehmm} outperform features learned via self-supervision on secondary structure and contact prediction, while learned features perform significantly better on remote homology detection.

Our results demonstrate that self-supervision is a promising paradigm for protein modeling but considerable improvements need to be made before self-supervised models can achieve breakthrough performance. All code and data for \benchmarkName are publically available\footnote{\url{https://github.com/songlab-cal/tape}}, and we encourage members of the machine learning community to participate in these exciting problems.


%% file: background.tex
\section{Background}
\label{sec:background}

\subsection{Protein Terminology}
\label{sec:protein-background}

Proteins are linear chains of amino acids connected by covalent bonds. We encode amino acids in the standard $25$-character alphabet, with $20$ characters for the standard amino acids, $2$ for the non-standard amino acids selenocysteine and pyrrolysine, $2$ for ambiguous amino acids, and $1$ for when the amino acid is unknown \cite{IUPACnomenclature, uniprot}. Throughout this paper, we represent a protein $x$ of length $L$ as a sequence of discrete amino acid characters $(x_1, x_2, \ldots, x_L)$ in this fixed alphabet.

Beyond its encoding as a sequence $(x_1, \ldots, x_L)$, a protein has a 3D molecular structure. 
The different levels of protein structure include 
\textit{primary} (amino acid sequence), \textit{secondary} (local features), and \textit{tertiary} (global features). Understanding how primary sequence folds into tertiary structure is a fundamental goal of biochemistry \cite{anfinsen}. Proteins are often made up of a few large \textit{protein domains}, sequences that are evolutionarily conserved,
and as such have a well-defined fold and function.

Evolutionary relationships between proteins arise because organisms must maintain certain functions, such as replicating DNA, as they evolve. Evolution has selected for proteins that are well-suited to these functions. Though structure is constrained by evolutionary pressures, sequence-level variation can be high, with very different sequences having similar structure \cite{creightonproteins}. Two proteins with different sequences but evolutionary or functional relationships are called \textit{homologs}.

Quantifying these evolutionary relationships is very important for preventing undesired information leakage between data splits. We mainly rely on \textit{sequence identity}, which measures the percentage of exact amino acid matches between aligned subsequences of proteins \cite{rost-sequence-alignment}. For example, filtering at a 25\% sequence identity threshold means that no two proteins in the training and test set have greater than 25\% exact amino acid matches. Other approaches besides sequence identity filtering also exist, depending on the generalization the task attempts to test \cite{brenner1998}.

\subsection{Modeling Evolutionary Relationships with Sequence Alignments}
\label{sec:modeling-evolutionary-relationships}

The key technique for modeling sequence relationships in computational biology is alignment \cite{blast, profilehmm, psi-blast, durbin}. Given a database of proteins and a new protein at test-time, an alignment-based method uses either carefully designed scoring systems to perform pairwise comparisons \cite{blast, smith-waterman}, Hidden Markov Model-like probabilistic models \cite{profilehmm}, or a combination \cite{psi-blast} to align the test protein against the database. If good alignments are found, information from the alignments is either directly sufficient for the task at hand, or can be fed into downstream models for further use \cite{arnold2006swiss}. 

\subsection{Semi-supervised Learning}
\label{sec:semi-supervised}

The fields of computer vision and natural language processing have been dealing with the question of how to learn from unlabeled data for years \cite{chapelle2009semi}. Images and text found on the internet generally lack accompanying annotations, yet still contain significant structure. Semi-supervised learning tries to jointly leverage information in the unlabeled and labeled data, with the goal of maximizing performance on the supervised task. One successful approach to learning from unlabeled examples is \textit{self-supervised learning}, which in NLP has taken the form of next-token prediction \cite{elmo}, masked-token prediction \cite{bert}, and next-sentence classification \cite{bert}. Analogously, there is good reason to believe that unlabelled protein sequences contain significant information about their structure and function \cite{anfinsen, altschuh}. Since proteins can be modeled as sequences of discrete tokens, we test both next-token and masked-token prediction for self-supervised learning.

%% file: related_work.tex
\section{Related Work}
\label{sec:related_work}

The most well-known protein modeling benchmark is the Critical Assessment of Structure Prediction (CASP) \cite{casp}, which focuses on structure modeling. Each time CASP is held, the test set consists of new experimentally validated structures which are held under embargo until the competition ends. This prevents information leakage and overfitting to the test set. The recently released ProteinNet \cite{proteinnet} provides easy to use, curated train/validation/test splits for machine learning researchers where test sets are taken from the CASP competition and sequence identity filtering is already performed. 
We take the contact prediction task from ProteinNet. However, we believe that structure prediction alone is not a sufficient benchmark for protein models, so we also use tasks not included in the CASP competition to give our benchmark a broader focus.


Semi-supervised learning for protein problems has been explored for decades, with lots of work on kernel-based pretraining \cite{protein-cluster-kernels, shin2006prediction}. These methods demonstrated that semi-supervised learning improved performance on protein network prediction and homolog detection, but couldn't scale beyond hundreds of thousands of unlabeled examples. Recent work in protein representation learning has proposed a variety of methods that apply NLP-based techniques for transfer learning to biological sequences \cite{bepler, unirep, seqvec, fair-bert}. In a related line of work, Riesselman \etal \cite{deepsequence} trained Variational Auto Encoders on aligned families of proteins to predict the functional impact of mutations. Alley \etal \cite{unirep} also try to combine self-supervision with alignment in their work by using alignment-based querying to build task-specific pretraining sets. 

Due to the relative infancy of protein representation learning as a field, the methods described above share few, if any, benchmarks. For example, both Rives \etal \cite{fair-bert} and Bepler \etal \cite{bepler} report transfer learning results on secondary structure prediction and contact prediction, but they differ significantly in test set creation and data-splitting strategies. Other self-supervised work such as Alley \etal \cite{unirep} and Yang \etal \cite{arnold-embedding} report protein engineering results, but on different tasks and datasets. 
With such varied task evaluation, it is challenging to assess the relative merits of different self-supervised modeling approaches, hindering efficient progress. 

%% file: datasets.tex
\section{Datasets}
\label{sec:datasets}

Here we describe our unsupervised pretraining and supervised benchmark datasets.  To create benchmarks that test generalization across large evolutionary distances and are useful in real-life scenarios, we curate specific training, validation, and test splits for each dataset. Producing the data for these tasks requires significant effort by experimentalists, database managers, and others. Following similar benchmarking efforts in NLP \cite{superglue}, we describe a set of citation guidelines in our repository\footnote{\url{https://github.com/songlab-cal/tape\#citation-guidelines}} to ensure these efforts are properly acknowledged.



\subsection{Unlabeled Sequence Dataset}
\label{sec:unlabeled}

We use Pfam \cite{pfam}, a database of thirty-one million protein domains used extensively in bioinformatics, as the pretraining corpus for \benchmarkName.  Sequences in Pfam are clustered into evolutionarily-related groups called \textit{families}. We leverage this structure by constructing a test set of fully heldout families (see Section \ref{sec:supplement-pfam-heldout} for details on the selected families), about 1\% of the data. For the remaining data we construct training and test sets using a random 95/5\% split. Perplexity on the uniform random split test set measures in-distribution generalization, while perplexity on the heldout families test set measures out-of-distribution generalization to proteins that are less evolutionarily related to the training set.

\subsection{Supervised Datasets}
\label{sec:supervised}

We provide five biologically relevant downstream prediction tasks to serve as benchmarks. We categorize these into structure prediction, evolutionary understanding, and protein engineering tasks. The datasets vary in size between 8 thousand and 50 thousand training examples (see Table \ref{tab:dataset-sizes} for sizes of all training, validation and test sets). Further information on data processing, splits and experimental challenges is in Appendix~\ref{sec:supplement-datasets}. For each task we provide:


\newcommand{\definition}{\textbf{(Definition)}}
\newcommand{\impact}{\textbf{(Impact)}}
\newcommand{\generalization}{\textbf{(Generalization)}}
\newcommand{\metric}{\textbf{(Metric)}}

\begin{itemize}[noitemsep, align=left]
    \item[\definition] A formal definition of the prediction problem, as well as the source of the data.
    \item[\impact] The impact of improving performance on this problem.
    \item[\generalization] The type of understanding and generalization desired.
    \item[\metric] The metric used in Table \ref{tab:downstream-results} to report results.
\end{itemize}

\begin{figure}
    \centering
    \subfloat[Secondary Structure]{
    \includegraphics[width=0.30\linewidth]{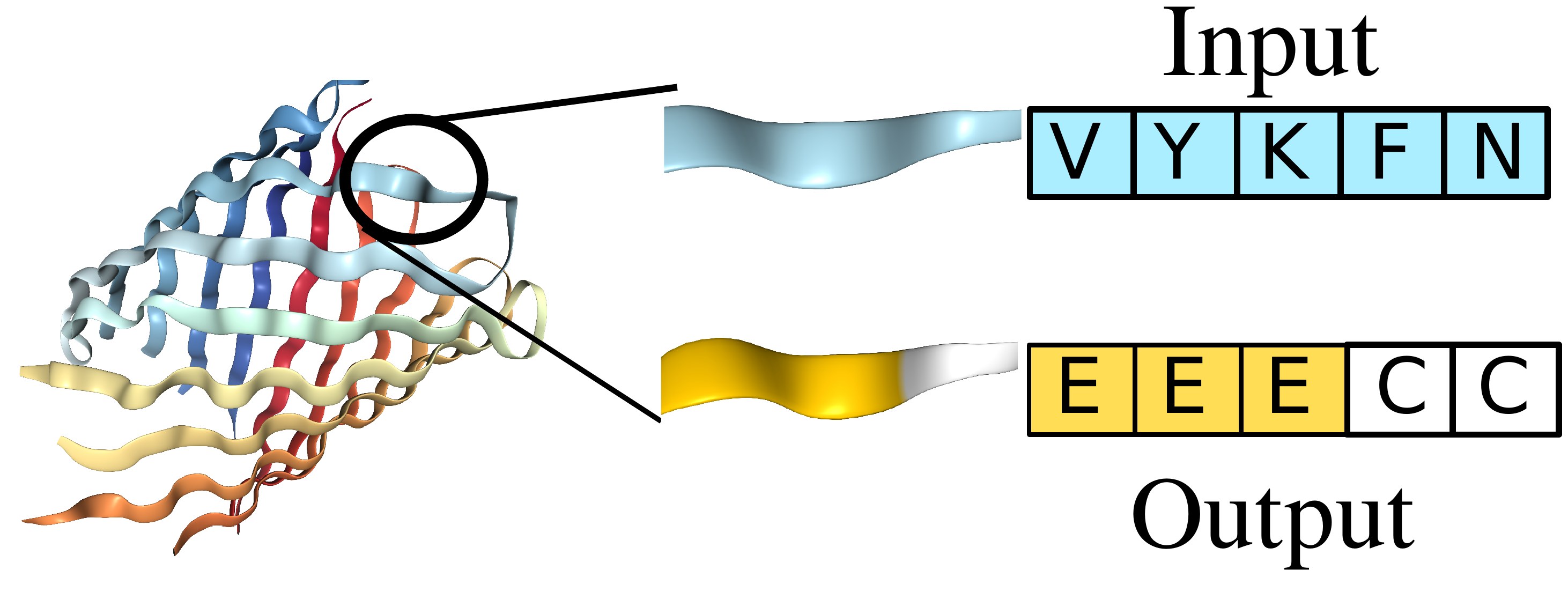}
    \label{fig:secondary-structure-task}
    }
    \hfill
    \subfloat[Contact Prediction]{
    \includegraphics[width=0.30\linewidth]{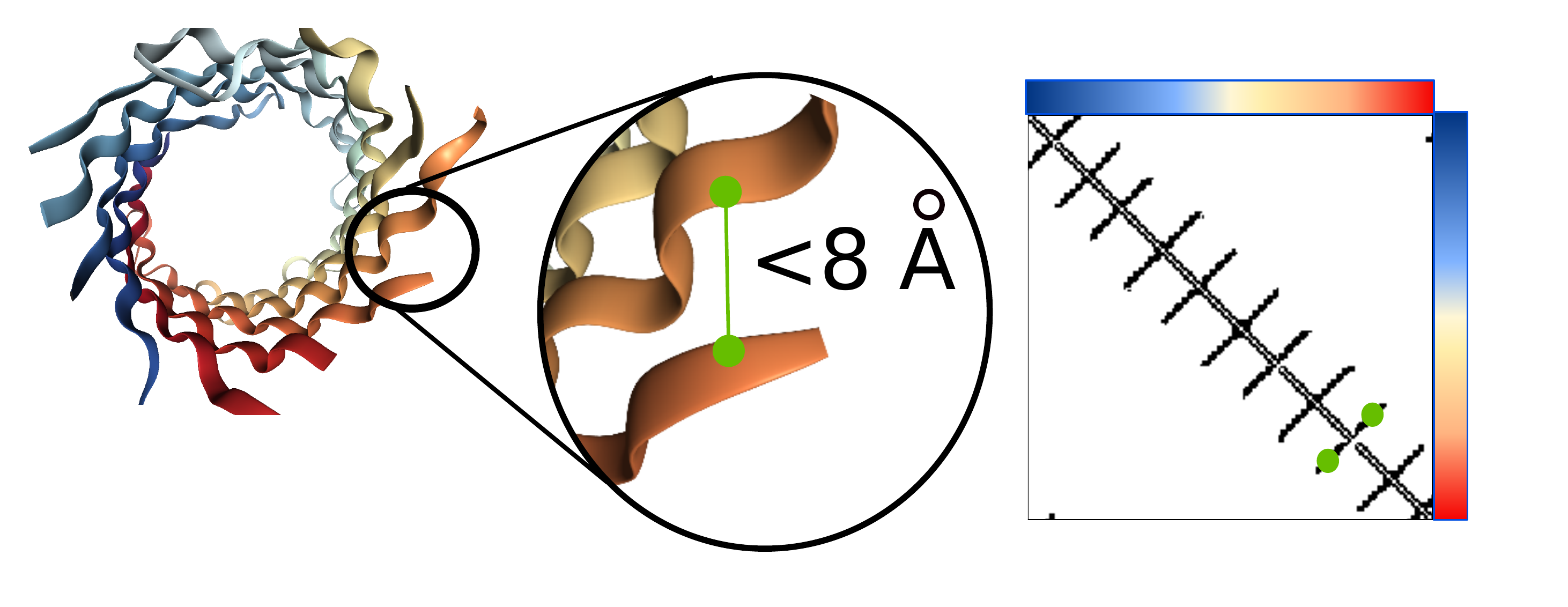}
    \label{fig:contact-prediction-task}
    }
    \hfill
    \subfloat[Remote Homology]{
    \includegraphics[width=0.30\linewidth]{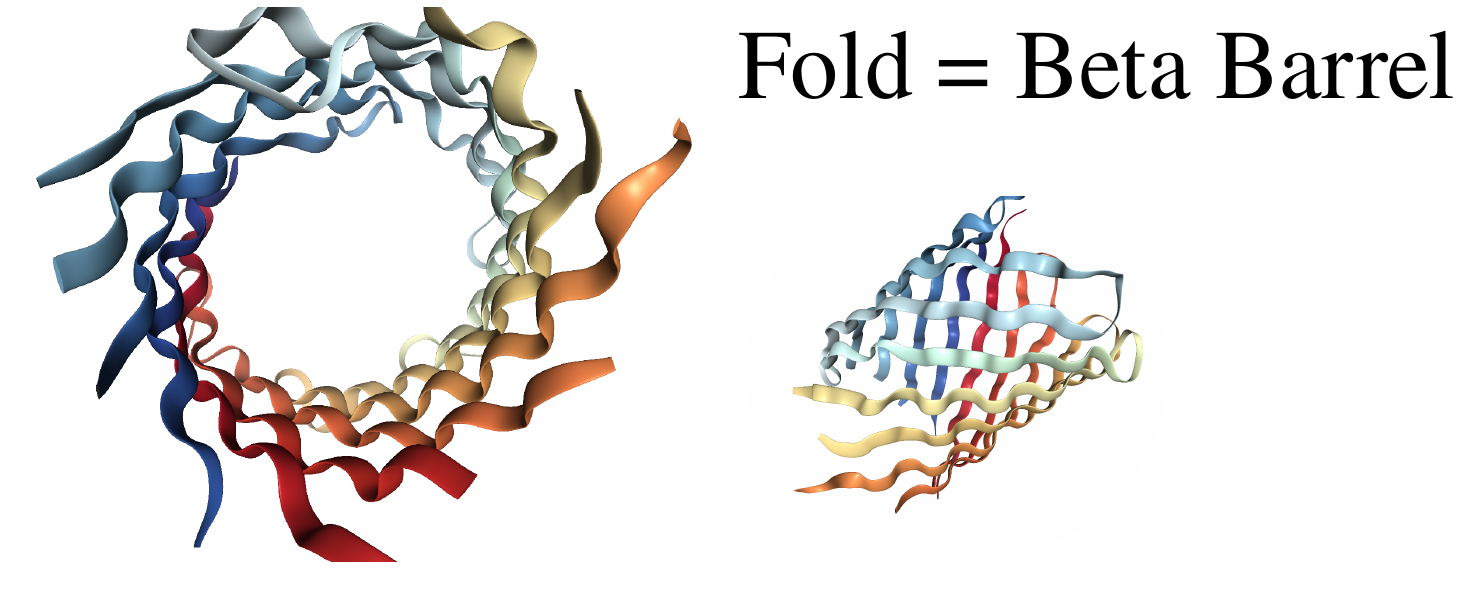}
    \label{fig:remote-homology-task}
    }
    \caption{Structure and Annotation Tasks on protein KgdM Porin (pdbid: 4FQE). (a) Viewing this Porin from the side, we show secondary structure, with the input amino acids for a segment (blue) and corresponding secondary structure labels (yellow and white). (b) Viewing this Porin from the front, we show a contact map, where entry $i$, $j$ in the matrix indicates whether amino acids at positions $i$, $j$ in the sequence are within 8 angstroms of each other. In green is a contact between two non-consecutive amino acids. (c) The fold-level remote homology class for this protein.}
    \label{fig:structure-and-evo-tasks}
\end{figure}

\paragraph{Task 1: Secondary Structure (SS) Prediction (Structure Prediction Task)}
\label{sec:secondary_structure}

\begin{itemize}[leftmargin=*, noitemsep, nolistsep, align=left]
    \item[\definition]  Secondary structure prediction is a sequence-to-sequence task where each input amino acid $x_i$ is mapped to a label $y_i \in \{\textrm{Helix}, \textrm{Strand}, \textrm{Other}\}$. See Figure \ref{fig:secondary-structure-task} for illustration. The data are from Klausen \etal \cite{netsurfp}.
    \item[\impact] SS is an important feature for understanding the function of a protein, especially if the protein of interest is not evolutionarily related to proteins with known structure \cite{netsurfp}. SS prediction tools are very commonly used to create richer input features for higher-level models \cite{jpred}.
    \item[\generalization] SS prediction tests the degree to which models learn local structure. Data splits are filtered at $25$\% sequence identity to test for broad generalization.
    \item[\metric] We report accuracy on a per-amino acid basis on the CB513 \cite{cb513} dataset.
\end{itemize}

\paragraph{Task 2: Contact Prediction (Structure Prediction Task)}
\label{sec:contact_prediction}

\begin{itemize}[leftmargin=*, noitemsep, nolistsep, align=left]
    \item[\definition] Contact prediction is a pairwise amino acid task, where each pair $x_i, x_j$ of input amino acids from sequence $x$ is mapped to a label $y_{ij} \in \{0, 1\}$, where the label denotes whether the amino acids are ``in contact'' (< 8\AA\, apart) or not. See Figure \ref{fig:contact-prediction-task} for illustration. The data are from the ProteinNet dataset \cite{proteinnet}.
    \item[\impact] Accurate contact maps provide powerful global information; e.g., they facilitate robust modeling of full 3D protein structure  \cite{kim}.  Of particular interest are medium- and long-range contacts, which may be as few as twelve sequence positions apart, or as many as hundreds apart.
    \item[\generalization] The abundance of medium- and long-range contacts makes contact prediction an ideal task for measuring a model's understanding of global protein context. We select the data splits that was  filtered at $30$\% sequence identity to test for broad generalization.
    \item[\metric] We report precision of the $L / 5$ most likely contacts for medium- and long-range contacts on the ProteinNet CASP12 test set, which is a standard metric reported in CASP \cite{casp}.
\end{itemize}

\paragraph{Task 3: Remote Homology Detection (Evolutionary Understanding Task)}
\label{sec:remote_homology}

\begin{itemize}[leftmargin=*, align=left, noitemsep, nolistsep]
    \item[\definition] This is a sequence classification task where each input protein $x$ is mapped to a label $y \in \{1, \ldots, 1195\}$, representing different possible protein folds. See Figure \ref{fig:remote-homology-task} for illustration. The data are from Hou \etal~\cite{deepsf}.
    \item[\impact] Detection of remote homologs is of great interest in microbiology and medicine; e.g., for detection of emerging antibiotic resistant genes \cite{amr-paper} and discovery of new CAS enzymes \cite{casx}.
    \item[\generalization] Remote homology detection measures a model's ability to detect structural similarity across distantly related inputs. We hold out entire evolutionary groups from the training set, forcing models to generalize across large evolutionary gaps.
    \item[\metric] We report overall classification accuracy on the fold-level heldout set from Hou \etal \cite{deepsf}.
\end{itemize}

\begin{figure}
    \centering
    \subfloat[Fluorescence]{
    \includegraphics[width=0.40\linewidth]{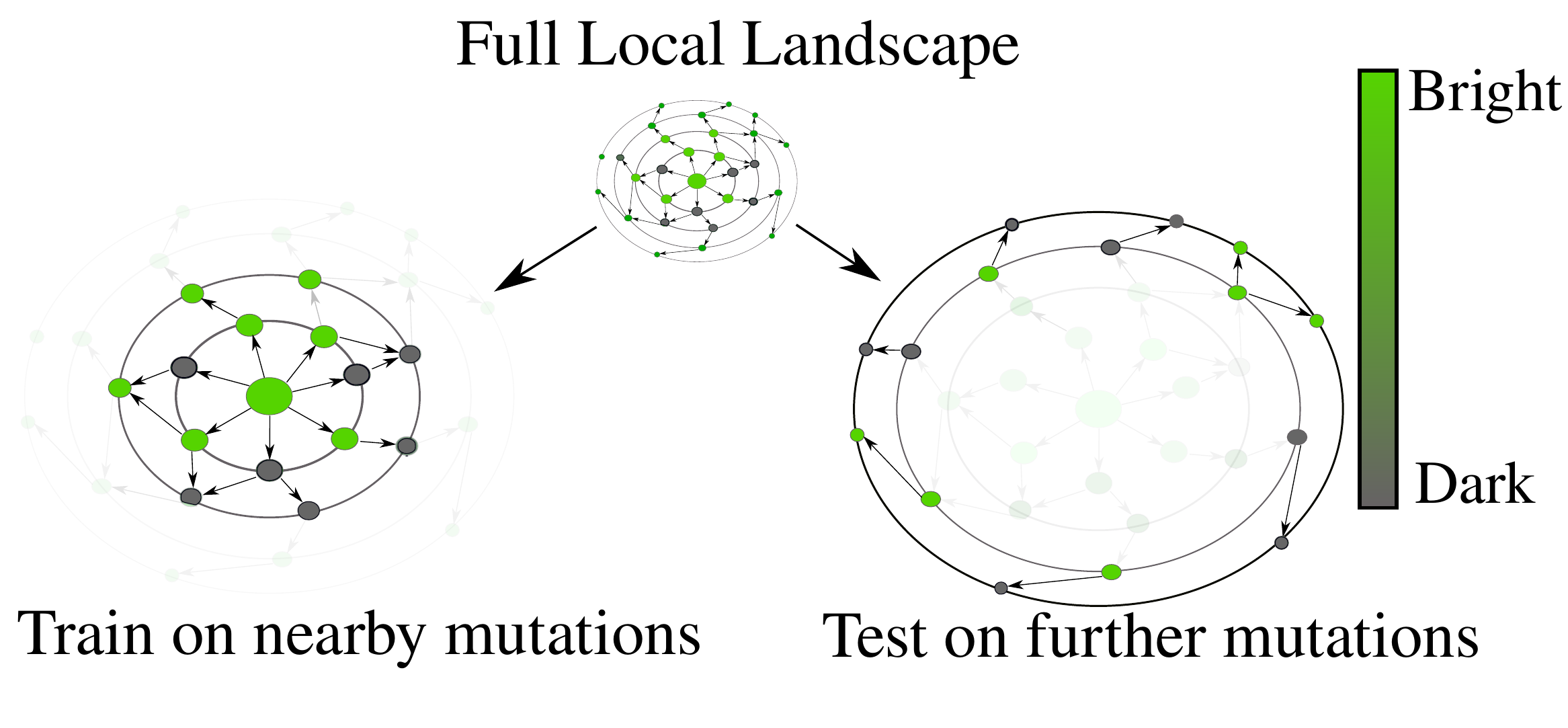}
    \label{fig:fluorescence-task}
    }
    \hfill
    \subfloat[Stability]{
    \includegraphics[width=0.49\linewidth]{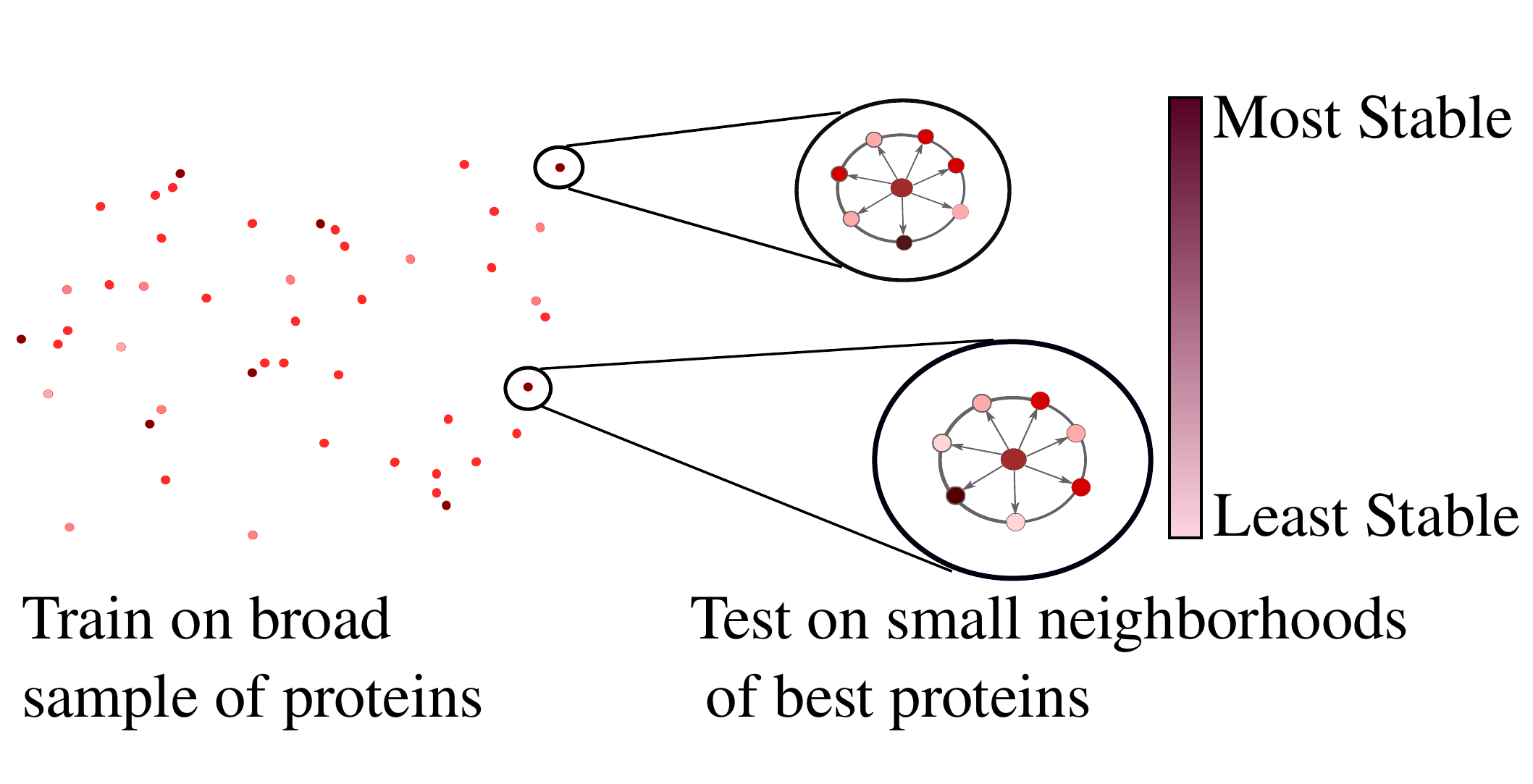}
    \label{fig:stability-task}
    }
    \caption{Protein Engineering Tasks. In both tasks, a parent protein $p$ is mutated to explore the local landscape. As such, dots represent proteins and directed arrow $x \to y$ denotes that $y$ has exactly one more mutation than $x$ away from  parent $p$. (a) The Fluorescence task consists of training on small neighborhood of the parent green fluorescent protein (GFP) and then testing on a more distant proteins. (b) The Stability task consists of training on a broad sample of proteins, followed by testing on one-mutation neighborhoods of the most promising sampled proteins.}
    \label{fig:protein-engineering-tasks}
\end{figure}

\paragraph{Task 4: Fluorescence Landscape Prediction (Protein Engineering Task)}
\label{sec:fluorescence}

\begin{itemize}[leftmargin=*, align=left, noitemsep, nolistsep]
    \item[\definition] This is a regression task where each input protein $x$ is mapped to a label $y \in \mathbb{R}$, corresponding to the log-fluorescence intensity of $x$. See Figure \ref{fig:fluorescence-task} for illustration. The data are from Sarkisyan \etal \cite{gfp-paper}.
    \item[\impact] For a protein of length $L$, the number of possible sequences $m$ mutations away is $O(L^m)$, a prohibitively large space for exhaustive search via experiment, even if $m$ is modest.  Moreover, due to epistasis (second- and higher-order interactions between mutations at different positions), greedy optimization approaches are unlikely to succeed.  Accurate computational predictions could allow significantly more efficient exploration of the landscape, resulting in better optima. Machine learning methods have already seen some success in related protein engineering tasks~\cite{arnold-ml-review}.
    \item[\generalization] The fluorescence prediction task tests the model's ability to distinguish between very similar inputs, as well as its ability to generalize to unseen combinations of mutations. The train set is a Hamming distance-$3$ neighborhood of the parent green fluorescent protein (GFP), while the test set has variants with four or more mutations.
    \item[\metric] We report Spearman's $\rho$ (rank correlation coefficient) on the test set.
\end{itemize}


\paragraph{Task 5: Stability Landscape Prediction (Protein Engineering Task)}
\label{sec:thermostability}

\begin{itemize}[leftmargin=*, align=left, noitemsep, nolistsep]
    \item[\definition] This is a regression task where each input protein $x$ is mapped to a label $y \in \mathbb{R}$ measuring the most extreme circumstances in which protein $x$ maintains its fold above a concentration threshold (a proxy for intrinsic stability). See Figure \ref{fig:stability-task} for illustration. The data are from Rocklin \etal \cite{stability-paper}.
    \item[\impact] Designing stable proteins is important to ensure, for example, that drugs are delivered before they are degraded. More generally, given a broad sample of protein measurements, finding better refinements of top candidates is useful for maximizing yield from expensive protein engineering experiments.
    \item[\generalization] This task tests a model's ability to generalize from a broad sampling of relevant sequences and to localize this information in a neighborhood of a few sequences, inverting the test-case for fluorescence above. The train set consists of proteins from four rounds of experimental design, while the test set contains 
    Hamming distance-$1$ neighbors of top candidate proteins. 
    \item[\metric] We report Spearman's $\rho$ on the test set.
\end{itemize}

%% file: models.tex
\section{Models and Experimental Setup}
\label{sec:models}

\begin{table}
    \caption{Language modeling metrics}
    \label{tab:language-modeling-metrics}
    \centering
    \begin{tabular}{lcccccc}
        \toprule
        \multicolumn{1}{c}{} & \multicolumn{3}{c}{Random Families} & \multicolumn{3}{c}{Heldout Families}\\
        \cmidrule(l){2-4} \cmidrule(l){5-7}
          & Accuracy & Perplexity & ECE & Accuracy & Perplexity & ECE\\
        \midrule
        \midrule
        Transformer                    & \bf{0.45} & \textbf{8.89} & \bf{6.01} & \textbf{0.30} & \textbf{13.04} & \textbf{10.04} \\
        LSTM                           & 0.40 & \textbf{8.89} & 6.94 & 0.16 & 14.72     & 15.21 \\
        ResNet                         & 0.41 & 10.16 & 6.86 & 0.29 & 13.55 & 10.32 \\
        Supervised LSTM \cite{bepler}      & 0.28 & 11.62 & 10.17 & 0.14 & 15.28 & 16.02 \\
        UniRep mLSTM \cite{unirep}            & 0.32 & 11.29 & 9.08 & 0.12 & 16.36 & 16.92 \\
        Random                         & 0.04 & 25 & 25 & 0.04 & 25 & 25 \\
        \bottomrule
    \end{tabular}
\end{table}

\paragraph{Losses:} We examine two self-supervised losses that have seen success in NLP. The first is \textit{next-token prediction} \cite{mikolov2010recurrent}, which models $p(x_i \mid x_1, \ldots, x_{i - 1})$. Since many protein tasks are sequence-to-sequence and require bidirectional context, we apply a variant of next-token prediction which additionally trains the reverse model, $p(x_i \mid x_{i + 1}, \ldots, x_L)$, providing full context at each position (assuming a Markov sequence). The second is \textit{masked-token prediction} \cite{bert}, which models\break $p(x_{\textmd{masked}} \mid x_{\textmd{unmasked}})$ by replacing the value of tokens at multiple positions with alternate tokens.

\paragraph{Protein-specific loss:} In addition to self-supervised algorithms, we explore another protein-specific training procedure proposed by Bepler \etal \cite{bepler}. They suggest that further \textit{supervised} pretraining of models can provide significant benefits. In particular, they propose supervised pretraining on contact prediction and remote homology detection, and show it increases performance on secondary structure prediction. Similar work in computer vision has shown that supervised pretraining can transfer well to other tasks, making this a promising avenue of exploration \cite{decaf}.

\paragraph{Architectures and Training:} We implement three  architectures: an LSTM \cite{lstm}, a Transformer \cite{vaswani}, and a dilated residual network (ResNet) \cite{dilated-residual-networks}. We use a 12-layer Transformer with a hidden size of 512 units and 8 attention heads, leading to a 38M-parameter model. Hyperparameters for the other models were chosen to approximately match the number of parameters in the Transformer. Our LSTM consists of two three-layer LSTMs with 1024 hidden units corresponding to the forward and backward language models, whose outputs are concatenated in the final layer, similar to ELMo \cite{elmo}. For the ResNet we use 35 residual blocks, each containing two convolutional layers with 256 filters, kernel size 9, and dilation rate 2.

In addition, we benchmark two previously proposed architectures that differ significantly from the three above. The first, proposed by Bepler \etal \cite{bepler}, is a two-layer bidirectional language model, similar to the LSTM discussed above, followed by three 512 hidden unit bidirectional LSTMs. The second, proposed by Alley \etal \cite{unirep}, is a unidirectional mLSTM \cite{mlstm} with 1900 hidden units. Details on implementing and training these architectures can be found in the original papers.

The Transformer and ResNet are trained with masked-token prediction, while the LSTM is trained with next-token prediction. Both Alley \etal and Bepler \etal are trained with next-token prediction. All self-supervised models are trained on four NVIDIA V100 GPUs for one week. 

\paragraph{Baselines:} We evaluate learned features against two baseline featurizations. The first is a one-hot encoding of the input amino acid sequence, which provides a simple baseline. Additionally, most current state-of-the-art algorithms for protein modeling take advantage of alignment or HMM-based inputs (see Section \ref{sec:modeling-evolutionary-relationships}). Alignments can be transformed into various features, such as mutation probabilities \cite{deepsf} or the HMM state-transition probabilities \cite{netsurfp} for each amino acid position. These are concatenated to the one-hot encoding of the amino acid to form another baseline featurization. For our baselines we use alignment-based inputs that vary per task depending on the inputs used by the current state-of-the-art method. See Appendix~\ref{sec:supplement-supervised-architectures} for details on the alignment-based features used. We do not use alignment-based inputs for protein engineering tasks - since proteins in the engineering datasets differ by only a single amino acid, and alignment-based methods search for proteins with high sequence identity, the alignment-based methods return the same set of features for all proteins we wish to distinguish between.

\paragraph{Experimental Setup:} The goal of our experimental setup is to systematically compare all featurizations. For each task we select a particular supervised architecture, drawing from state-of-the-art where available, and make sure that fine-tuning on all language models is identical. See Appendix~\ref{sec:supplement-supervised-architectures} for details on supervised architectures and training.

%% file: results.tex
\section{Results}
\label{sec:results}


Table \ref{tab:language-modeling-metrics} contains accuracy, perplexity, and exponentiated cross entropy (ECE) on the language modeling task for the five architectures we trained with self-supervision as well as a random model baseline.  We report metrics on both the random split and the fully heldout families. Supervised LSTM metrics are reported after language modeling pretraining, but before supervised pretraining. Heldout family accuracy is consistently lower than random-split accuracy, demonstrating a drop in the out-of-distribution generalization ability. Note that although some models have lower perplexity than others on both random-split and heldout sets, this lower perplexity does not necessarily correspond to better performance on downstream tasks. This replicates the finding in Rives \etal \cite{fair-bert}. 

\begin{table}
    \caption{Results on downstream supervised tasks 
    }
    \label{tab:downstream-results}
    \centering
    \begin{tabular}{llccccc}
        \toprule
        \multicolumn{2}{c}{Method} & \multicolumn{2}{c}{Structure} & \multicolumn{1}{c}{Evolutionary} & \multicolumn{2}{c}{Engineering}\\
        \cmidrule(l){3-4} \cmidrule(l){5-5} \cmidrule(l){6-7}
          & & SS & Contact & Homology & Fluorescence & Stability \\
        \midrule
        \multirow{3}*{No Pretrain} & Transformer & 0.70 & 0.32 & 0.09 & 0.22 & -0.06 \\
                                   & LSTM        & 0.71 & 0.19 & 0.12 & 0.21 & 0.28 \\
                                   & ResNet      & 0.70 & 0.20 & 0.10 & -0.28 & 0.61 \\
                                   \cmidrule(l){2-7}
        \multirow{3}*{Pretrain}    & Transformer & 0.73 & 0.36 & 0.21 &\textbf{0.68} & \textbf{0.73} \\
                                   & LSTM        & 0.75 & 0.39 & \textbf{0.26} & 0.67 & 0.69 \\
                                   & ResNet      & 0.75 & 0.29 & 0.17 & 0.21 & \textbf{0.73} \\
                                   \cmidrule(l){2-7}
        Supervised \cite{bepler}  & LSTM         & 0.73 & 0.40 & 0.17 & 0.33 & 0.64 \\
        UniRep   \cite{unirep} & mLSTM           & 0.73 & 0.34 & 0.23 & 0.67 & \textbf{0.73} \\
                                    \cmidrule(l){2-7}
        \multirow{2}*{Baseline}    & One-hot      & 0.69 & 0.29 & 0.09 & 0.14 & 0.19 \\
                                   & Alignment & \textbf{0.80} & \textbf{0.64} & 0.09 & N/A & N/A \\
        \bottomrule
    \end{tabular}
\end{table}



\begin{figure}
    \centering
    \subfloat[Dark Mode]{
    \includegraphics[width=0.32\linewidth]{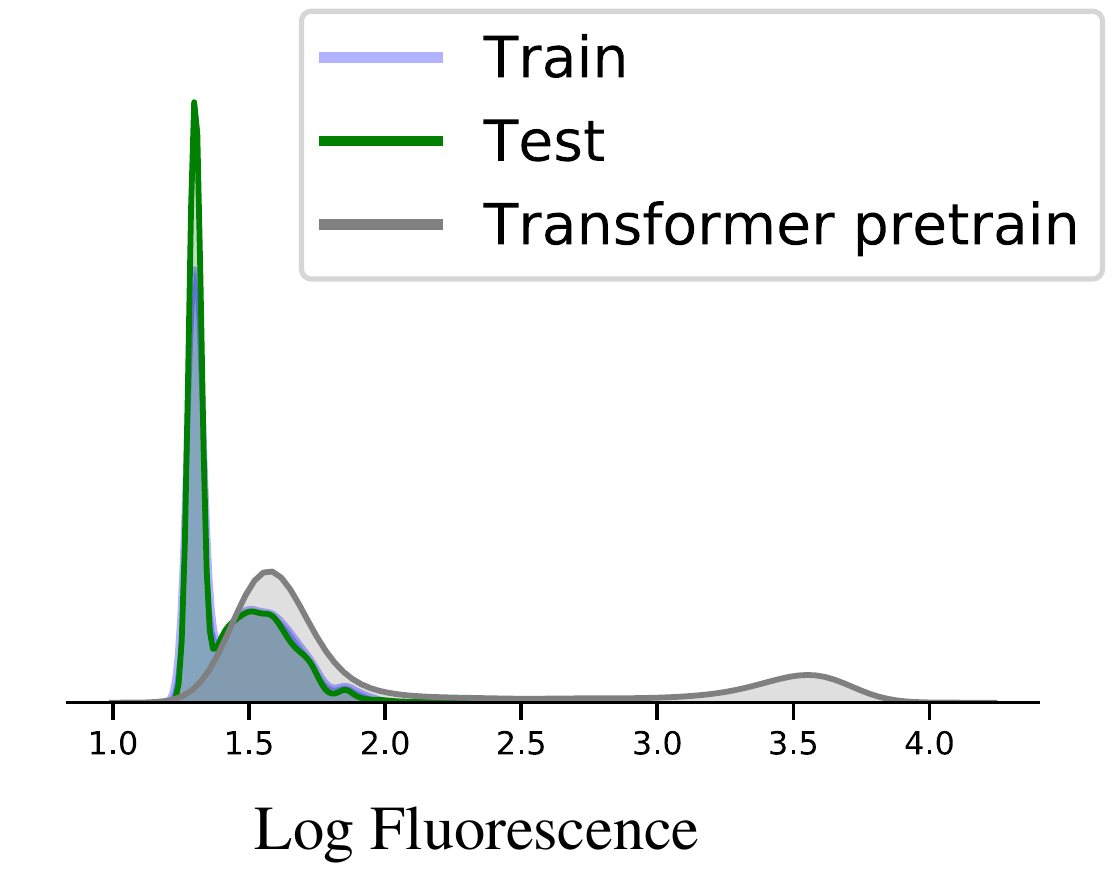}
    }
    \hfill
    \subfloat[Bright Mode]{
    \includegraphics[width=0.32\linewidth]{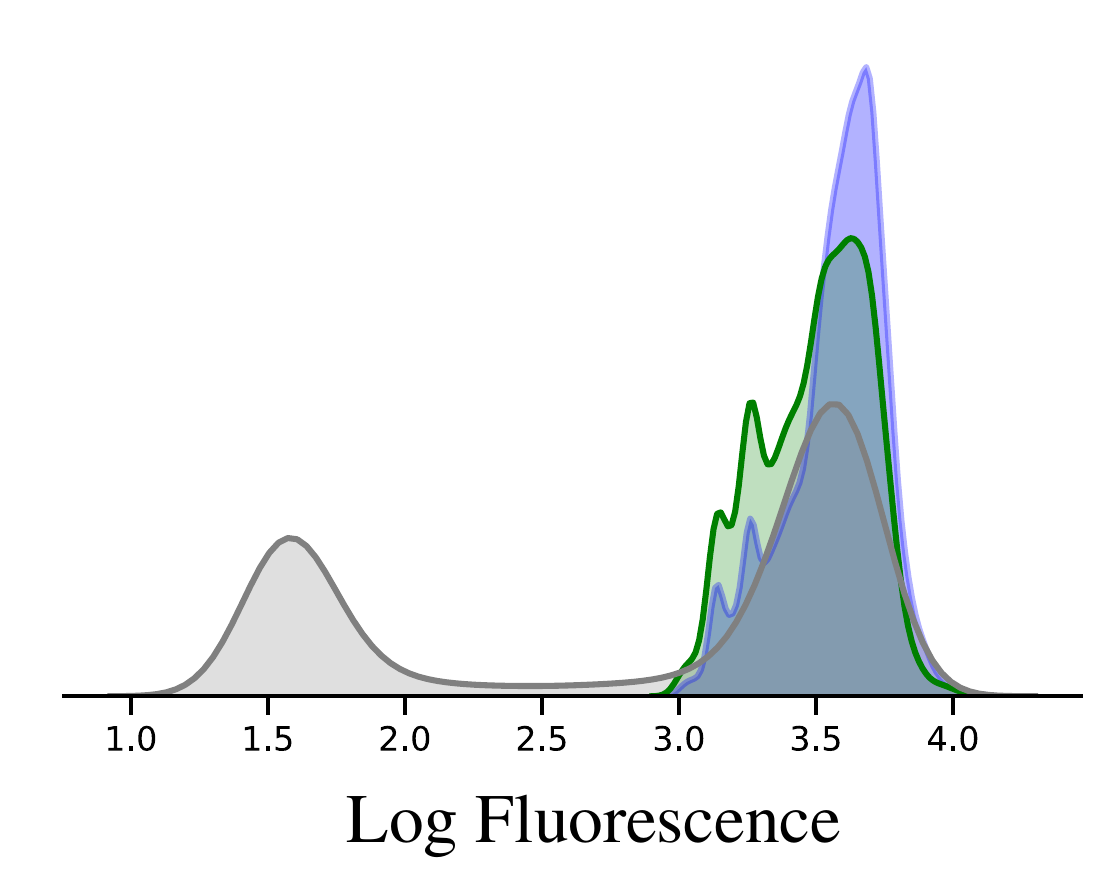}
    }
    \hfill
    \subfloat[Embedding t-SNE]{
    \includegraphics[width=0.32\linewidth]{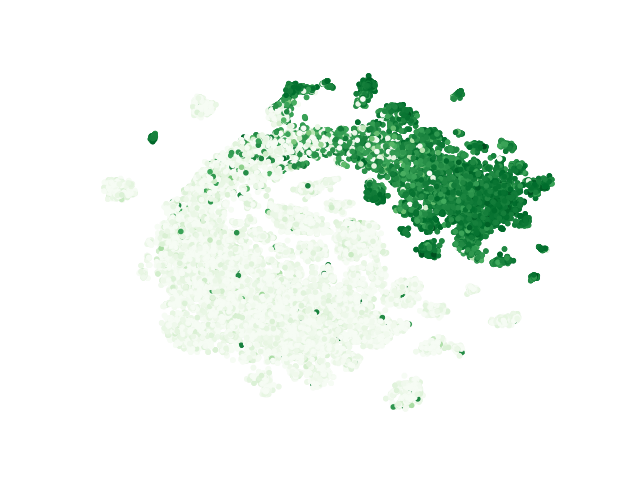}
    }
    \caption{Distribution of training, test, and pretrained Transformer predictions on the dark and bright modes, along with t-SNE of pretrained Transformer protein embeddings colored by log-fluorescence.}
    \label{fig:fluorescence-analysis}
\end{figure}

Table \ref{tab:downstream-results} contains results for all benchmarked architectures and training procedures on all downstream tasks in \benchmarkName. We report accuracy, precision, or Spearman's $\rho$, depending on the task, so higher is always better and each metric has a maximum value of $1.0$. See Section \ref{sec:datasets} for the metric reported in each task. Detailed results and metrics for each task are in Appendix~\ref{sec:supplement-supervised-results}. 

We see from Table \ref{tab:downstream-results} that self-supervised pretraining improves overall performance across almost all models and all tasks. Further analysis reveals aspects of these tasks with more for significant improvement. In the fluorescence task, the distribution is bimodal with a mode of bright proteins and a mode of dark proteins (see Figure \ref{fig:fluorescence-analysis}). Since one goal of using machine learning models in protein engineering is to screen potential variants, it is important for these methods to successfully distinguish between beneficial and deleterious mutations. Figure \ref{fig:fluorescence-analysis} shows that the model does successfully perform some clustering of fluorescent proteins, but that many proteins are still misclassified.

For the stability task, to identify which mutations a model believes are beneficial, we use the parent protein as a decision boundary and label a mutation as beneficial if its predicted stability is higher than the parent's predicted stability. We find that our  best pretrained model achieves 70\% accuracy in making this prediction while our best non-pretrained model achieves 68\% accuracy (see Table \ref{tab:stability-full-dataset} for full results). Improving the ability to distinguish beneficial from deleterious mutations would make these models much more useful in real protein engineering experiments.

\begin{figure}
    \centering
    \subfloat[True Contacts]{
    \includegraphics[width=0.19\linewidth]{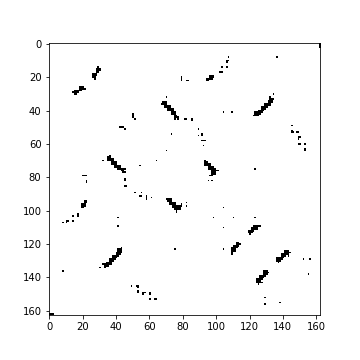}
    }
    \subfloat[LSTM]{
    \includegraphics[width=0.19\linewidth]{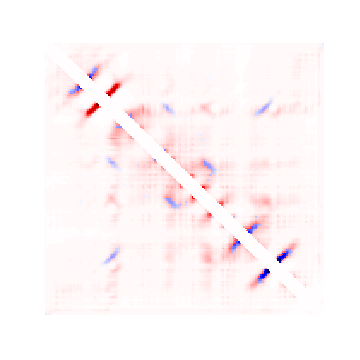}
    }
    \subfloat[LSTM Pretrain]{
    \includegraphics[width=0.19\linewidth]{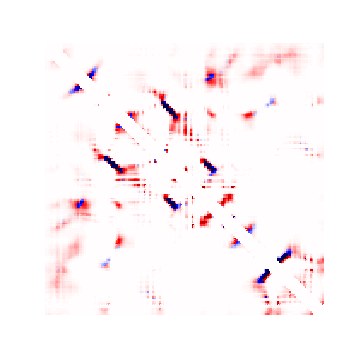}
    }
    \subfloat[One Hot]{
    \includegraphics[width=0.19\linewidth]{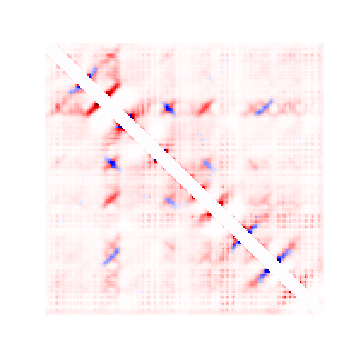}
    }
    \subfloat[Alignment]{
    \includegraphics[width=0.19\linewidth]{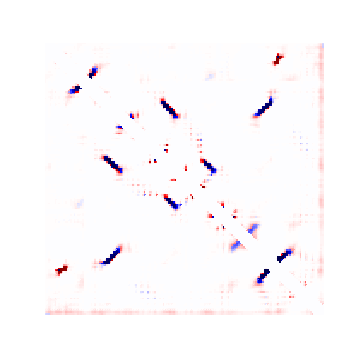}
    }
    \caption{Predicted contacts for chain 1A of a Bacterioferritin comigratory protein (pdbid: 3GKN). Blue indicates true positive contacts while red indicates false positive contacts. Darker colors represent more certainty from the model.}
    \label{fig:contact-analysis}
\end{figure}

In the contact prediction task, long-range contacts are of particular interest and can be hundreds of positions apart. Figure \ref{fig:contact-analysis} shows the predictions of several models on a protein where the longest range contact occurs between the 8th and 136th amino acids. Pretraining helps the model capture more long-range information and improves the overall resolution of the predicted map. However, the hand-engineered alignment features result in a much sharper map, accurately resolving a smaller number well-spaced of long-range contacts. This increased specificity is highly relevant in the structure prediction pipeline \cite{casp12-contact, kim} and highlights a clear challenge for pretraining.

%% file: discussion.tex
\section{Discussion}
\label{sec:discussion} 

\paragraph{Comparison to state of the art.}
As shown in Table \ref{tab:downstream-results}, alignment-based inputs can provide a powerful signal that outperforms current self-supervised models on multiple tasks. Current state-of-the-art prediction methods for secondary structure prediction, contact prediction, and remote homology classification all take in alignment-based inputs. These methods combine alignment-based inputs with other techniques (e.g. multi-task training, kernel regularization) to achieve an additional boost in performance. For comparison, NetSurfP-2.0 \cite{netsurfp} achieves 85\% accuracy on the CB513 \cite{cb513} secondary structure dataset, compared to our best model's 75\% accuracy, RaptorX \cite{raptorx} achieves 0.69 precision at $L / 5$ on CASP12 contact prediction, compared to our best model's 0.49, and DeepSF \cite{deepsf} achieves 41\% accuracy on remote homology detection compared to our best model's 26\%.

\paragraph{Need for multiple benchmark tasks.}
Our results support our hypothesis that multiple tasks are required to appropriately benchmark performance of a given method. Our Transformer, which performs worst of the three models in secondary structure prediction, performs best on the fluorescence and stability tasks. The reverse is true of our ResNet, which ties the LSTM in secondary structure prediction but performs far worse for the fluorescence task, with a Spearman's $\rho$ of $0.21$ compared to the LSTM's $0.67$. This shows that performance on a single task does not capture the full extent of a trained model's knowledge and biases, creating the need for multi-task benchmarks such as \benchmarkName.

%% file: future_work.tex
\section{Future Work}
\label{sec:future-work}

Protein representation learning is an exciting field with incredible room for expansion, innovation, and impact. The exponentially growing gap between labeled and unlabeled protein data means that self-supervised learning will continue to play a large role in the future of computational protein modeling. Our results show that no single self-supervised model performs best across all protein tasks. We believe this is a clear challenge for further research in self-supervised learning, as there is a huge space of model architecture, training procedures, and unsupervised task choices left to explore. It may be that language modeling as a task is not enough, and that protein-specific tasks are necessary to push performance past state of the art. Further exploring the relationship between alignment-based and learned representations will be necessary to capitalize on the advantages of each. We hope that the datasets and benchmarks in \benchmarkName will provide a systematic model-evaluation framework that allows more machine learning researchers to contribute to this field. 


%% file: acknowledgements.tex
\paragraph{Acknowledgments} 
We thank Philippe Laban, David Chan, Jeffrey Spence, Jacob West-Roberts, Alex Crits-Cristoph, Aravind Srinivas, Surojit Biswas, Ethan Alley, Mohammed AlQuraishi and Grigory Khimulya for valuable input on this paper. We thank the AWS Educate program for providing us with the resources to train our models. Additionally, we acknowledge funding from Berkeley Deep Drive, Chan-Zuckerberg Biohub, DARPA XAI, NIH, the Packard Fellowship for Science and Engineering, and the Open Philanthropy Project.

%% file: supplement.tex
\appendix
\section{Appendix}
\beginsupplement
\label{sec:supplement}

\newcommand{\dataset}{\textbf{(Dataset)}}
\newcommand{\labeling}{\textbf{(Labeling)}}

\subsection{Dataset Details}
\label{sec:supplement-datasets}

\begin{table}
    \caption{Dataset sizes}
    \label{tab:dataset-sizes}
    \centering
    \begin{tabular}{llll}
        \toprule
        Task & Train & Valid & Test \\
        \midrule
        Language Modeling   & 32,207,059 &  N/A     & 2,147,130 (Random-split) / 44,314 (Heldout families) \\
        Secondary Structure & 8,678  & 2,170 & 513 (CB513) / 115 (TS115) / 21 (CASP12) \\
        Contact Prediction  & 25,299 & 224   & 40 (CASP12) \\
        Remote Homology     & 12,312 & 736   & 718 (Fold) / 1,254 (Superfamily) / 1,272 (Family)\\
        Fluorescence        & 21,446 & 5,362 & 27,217\\
        Stability     & 53,679 & 2,447 & 12,839 \\
        \bottomrule
    \end{tabular}
\end{table}

In Table \ref{tab:dataset-sizes} we show the size of all train, validation, and test sets.

We provide further details about dataset sources, preprocessing decisions, data splitting, and experimental challenges in obtaining labels for each of our supervised tasks below. For ease of reading, each section starts with the following items:
\begin{itemize}[leftmargin=*, align=left, noitemsep, nolistsep]
    \item[\dataset] The source of the dataset and creation of train/test splits.
    \item[\labeling] The current approach to acquiring supervised labels for this task.
\end{itemize}

\subsubsection{Secondary Structure Details}
\label{sec:supplement-secondary-structure-details}
\begin{itemize}[leftmargin=*, align=left, noitemsep, nolistsep]
    \item[\dataset]  We use a training and validation set from Klausen \etal \cite{netsurfp}, which is filtered such that no two proteins have greater than 25\% sequence identity. We use three test sets, CB513 \cite{cb513}, CASP12 \cite{casp12}, and TS115 \cite{ts115}. The training set is also filtered at the 25\% identity threshold with these test sets. This filtering tests the model's ability to generalize in the interesting case where test proteins are not closely related to train proteins.
    \item[\labeling] Determining the secondary structure of a protein experimentally requires high-resolution imaging of the structure, a particularly labor intensive task for structural biologists. Imaging often uses Cryo Electron-Microscopy or X-Ray Crystallography, which can take between weeks and years and can cost over \$$200,000$ \cite{structure-cost}.
\end{itemize}

\subsubsection{Contact Prediction Details}
\label{sec:supplement-contact-prediction-details}
\begin{itemize}[leftmargin=*, align=left, noitemsep, nolistsep]
    \item[\dataset] We use training, validation, and test sets from ProteinNet \cite{proteinnet}, which uses a test set based on the CASP12 \cite{casp12} competition, with training and validation sets filtered at the 30\% sequence identity threshold. This tests the ability of the model to generalize to proteins that are not closely related to any train proteins.
    \item[\labeling] Determining the contacts of a protein requires knowing its full 3D structure. As with secondary structure, determining the 3D structure requires imaging a protein.
\end{itemize}

\subsubsection{Remote Homology Details}
\label{sec:supplement-remote-homology-details}

\begin{itemize}[leftmargin=*, align=left, noitemsep, nolistsep]
    \item[\dataset] We use a training, validation, and test set from \cite{deepsf}, derived from the SCOP 1.75 database \cite{scope} of hierarchically classified protein domains. All proteins of a given fold are further categorized into related \textit{superfamilies}. Entire superfamilies are held out from the training set, allowing us to evaluate how the model generalizes across evolutionary distance when structure is preserved.
    \item[\labeling] Each fold is annotated from the structure of the sequence, which SCOP pulls from the Protein DataBank \cite{pdb, scope}. Finding new superfamilies with the same fold is a challenging task, requiring sequencing in extreme environments as is often done in metagenomics \cite{new-bacterial-domains}.
\end{itemize}

\subsubsection{Fluorescence Details}
\label{sec:supplement-fluorescence-details}
\begin{itemize}[leftmargin=*, align=left, noitemsep, nolistsep]
    \item[\dataset] We use data generated by an experimental technique called Deep Mutational Scanning (DMS) \cite{gfp-paper}. This technique allows for extensive characterizations of small neighborhoods of a parent protein through mutagenesis. We create training, validation, and test splits ourselves, partitioning the data so that train and validation are in a Hamming distance 3 neighborhood of the original protein, while test data is a sample from the Hamming distance 4-15 neighborhood.
    \item[\labeling] DMS is efficient for local characterization near a single protein, but its samples become vanishingly small once neighborhoods start to expand outside of Hamming distance $2$. 
\end{itemize}

\subsubsection{Stability Details}
\label{sec:supplement-thermostability-details}
\begin{itemize}[leftmargin=*, align=left, noitemsep, nolistsep]
    \item[\dataset] We use data generated by a novel combination of parallel DNA synthesis and protein stability measurements \cite{stability-paper}. We create training, validation, and test splits ourselves, partitioning the data so that training and validation sets come from four rounds of experimental data measuring stability for many candidate proteins, while our test set consists of seventeen 1-Hamming distance neighborhoods around promising proteins observed in the four rounds of experimentation.
    \item[\labeling] This approach for observing stability is powerful because of its throughput, allowing the authors to find the most stable proteins ever observed for certain classes \cite{stability-paper}. The authors observe that the computational methods used to guide their selection at each stage could be improved, meaning that in this case better models could actually lead to better labeled data in a virtuous cycle.
\end{itemize}

\subsection{Supervised Architectures}
\label{sec:supplement-supervised-architectures}

For each task, we fixed one supervised architecture and tried one-hot, alignment-based, and neural net based features. We did not perform hyperparameter tuning or significant architecture optimization, as the main goal was to compare feature extraction techniques.

For each task we define the supervised architecture below. If this is a state of the art architecture from other work, we highlight any novel training procedure or inputs they take.

\subsubsection{Secondary Structure Architecture}
\label{sec:supplement-secondary-structure-architecture}
We used the NetSurfP2.0 model from Klausen et al \cite{netsurfp}. The model consists of two convolutional layers followed by two bidirection LSTM layers and a linear output layer. The convolutional layers have filter size 32 and kernel size 129 and 257, respectively. The bidirectional LSTM layers have 1024 hidden units each. 

Our evolutionary features for secondary structure prediction are transition probabilities and state information from HHblits \cite{hhblits}, an HMM-HMM alignment method. In the original model, the authors take HHblits outputs in addition to a one-hot encoding of the sequence, giving $50$-dimensional inputs at each position. They train the model on multiple tasks including secondary structure prediction (3 and 8 class), bond-angle prediction, and solvent accessibility prediction. For clarity, we only compared to the model trained without the multitask training, which in our experiments contributed an extra one to two percent in test accuracy. In addition to multitask training, they balance the losses between different tasks to achieve maximum accuracy on secondary structure prediction. All features and code to do the full multitask training is available in our repository.

\subsubsection{Contact Prediction Architecture}
\label{sec:supplement-contact-prediction-architecture}
We used a supervised network inspired by the RaptorX-Contact model from Ma et al \cite{raptorx}. Since a contact map is a 2D pairwise prediction, we form a 2D input from our 1D features by concatenating the features at position $i$ and $j$ for all $i, j$. This 2D input is then passed through a convolutional residual network with. The 2D network contains 30 residual blocks with two convolutional layers each. Each convolution in the residual block has filter size 64 and a kernel size of 3.

Currently our evolutionary features for contact prediction are Position Specific Scoring Matrices (PSSMs) available in ProteinNet \cite{proteinnet}. The original RaptorX method has a more complex evolutionary featurization: they construct a Multiple Sequence Alignment for each protein, then pass it through CCMpred \cite{ccmpred} - a Markov Random Field based contact prediction method. This outputs a 2D featurization including mutual information and pairwise potential. This, along with 1D HMM alignment features and the one-hot encoding of each amino acid are fed to their network. We are currently recreating this pipeline to use these features instead of PSSMs, as the results reported by RaptorX are better than those with the PSSMs.

\subsubsection{Remote Homology Architecture}
\label{sec:remote-homology-architecture}
Remote homology requires a single prediction for each protein. To obtain a sequence-length invariant protein embedding we compute an attention-weighted mean of the amino acid embeddings. More precisely, we predict an attention value for each position in the sequence using a trainable dense layer, then use those attention values to compute an attention-weighted mean protein embedding. This protein embedding is then passed through a 512 hidden unit dense layer, a relu nonlinearity, and a final linear output layer to predict logits for all 1195 classes. We note that Hou \etal \cite{deepsf} propose a deep architecture for this task and report state of the art performance. When we compared the performance of this supervised architecture to that of the attention-weighted mean above, the attention-based embedding performed better for all featurizations. As such, we choose to report results using the simpler attention-based downstream architecture.

Our evolutionary features for remote homology detection are Position Specific Scoring Matrices (PSSMs), following the recent work DeepSF \cite{deepsf}. The current state of the art method in this problem, DeepSF \cite{deepsf}, takes in a one-hot encoding of the amino acids, predicted secondary structure labels, predicted solvent accessibility labels, and a 1D alignment-based features. In an ablation study, the authors show that the secondary structure labels are most useful for performance of their model. We report only one-hot and alignment-based results in the main paper to maintain consistency with alignment-based featurizations for other tasks. All input features used by DeepSF are available in our repository.

\subsubsection{Protein Engineering Architectures}
\label{sec:protein-engineering-architecture}
Protein engineering also requires a single prediction for each protein. Therefore we use the same architecture as we do for remote homology, computing an attention-weighted mean protein embedding, a dense layer with 512 hidden units, a relu nonlinearity and a final linear output layer to predict the quantity of interest (either stability or fluorescence).

Since we create these training, validation, and test splits ourselves, no clear previous state of the art exists. Related work on protein engineering has used a similar architecture by computing a single protein embedding followed by some form of projection (linear or with a small feed forward network) \cite{unirep, fair-bert}. These methods also do not take in alignment-based features and only use one-hot amino acids as inputs.

\subsection{Training Details}
\label{sec:supplement-training-details}

Self-supervised models were all trained on four NVIDIA V100 GPUs on AWS for 1 week. Training used a learning rate of $10^{-3}$ with a linear warm up schedule, the Adam optimizer, and a 10\% dropout rate. Since proteins vary in length significantly, we use variable batch sizes depending on the length of the protein. These sizes also differ based on model architecture, as some models (e.g. the Transformer) have significantly higher memory requirements. Specific batch sizes for each model at each protein length are available in our repository.

Supervised models were trained on two NVIDIA Titan Xp GPUs until convergence (no increase in validation accuracy for $10$ epochs) with the exception of the memory-intensive Contact Prediction task, which was trained on two NVIDIA Titan RTX GPUs until convergence. Training used a learning rate of $10^{-4}$ with a linear warm up schedule, the Adam optimizer, and a 10\% dropout rate. We backpropagated fully through all models during supervised fine-tuning.

In addition, due to high memory requirements of some downstream tasks (especially contact prediction) we use memory saving gradients \cite{memory-saving-gradients} to fit more examples per batch on the GPU.

\subsection{Pfam Heldout Families}
\label{sec:supplement-pfam-heldout}

The following Pfam clans were held out during self-supervised training: CL0635, CL0624, CL0355, CL0100, CL0417, CL0630. The following Pfam families were held out during self-supervised training: PF18346, PF14604, PF18697, PF03577, PF01112, PF03417. First, a ``clan'' is a cluster of families grouped by the maintainers of Pfam based on shared function or evolutionary origin (see \cite{pfam} for details). We chose holdout clans and families in pairs, where a clan of novel function is held out together with a family that is similar in sequence but different evolutionarily or functionally. This serves to simultaneously test generalization across large distances (entirely held out families) and between similar looking unseen groups (e.g. the paired holdout clan and holdout family).

\subsection{Bepler Supervised Training}
\label{sec:supplement-bepler-training}
We perform supervised pretraining using the same architecture described in Bepler \etal \cite{bepler}. We train on the same tasks, a paired remote homology task and contact map prediction task. However, in order to accurately report results on downstream secondary structure, contact map, and remote homology datasets, which were filtered by sequence identity, we perform this same sequence identity filtering on the supervised pretraining set. This reduced the supervised pretraining dataset size by 75\% which likely reduced the effectiveness of the supervised pretraining. Both filtered and unfiltered supervised pretraining datasets are made available in our repository.

\subsection{Detailed Results on Supervised Tasks}
\label{sec:supplement-supervised-results}

Here we provide detailed results on each task, examining multiple metrics and test-conditions to further determine what the models are learning. 

\subsubsection{Secondary Structure Results}
\label{sec:supplement-secondary-structure-results}

\begin{table}
    \caption{Detailed secondary structure results}
    \label{tab:secondary-structure}
    \centering
    \begin{tabular}{llcccccc}
        \toprule
        \multicolumn{2}{c}{} & \multicolumn{3}{c}{Three-Way Accuracy (Q3)} & \multicolumn{3}{c}{Eight-Way Accuracy (Q8)}\\
        \cmidrule(l){3-5} \cmidrule(l){6-8}
          & & CB513 & CASP12 & TS115 & CB513 & CASP12 & TS115\\
        \midrule
        \midrule
        \multirow{3}*{No Pretrain} & Transformer & 0.70 & 0.68 & 0.73 & 0.51 & 0.52 & 0.58 \\
                                   & LSTM        & 0.71 & 0.69 & 0.74 & 0.47 & 0.48 & 0.52 \\
                                   & ResNet      & 0.70 & 0.68 & 0.73 & 0.55 & 0.56 & 0.61\\
                                   \cmidrule(l){2-8}
        \multirow{3}*{Pretrain}    & Transformer & 0.73 & 0.71 & 0.77 & 0.59 & 0.59 & 0.64\\
                                   & LSTM        & 0.75 & 0.70 & 0.78 & 0.59 & 0.57 & 0.66\\
                                   & ResNet      & 0.75 & 0.72 & 0.78 & 0.58 & 0.58 & 0.64\\
                                   \cmidrule(l){2-8}
    Supervised \cite{bepler}       & LSTM        & 0.73 & 0.70 & 0.76 & 0.58 & 0.57 & 0.65 \\
    UniRep \cite{unirep}           & mLSTM       & 0.73 & 0.72 & 0.77 & 0.57 & 0.59 & 0.63\\
                                    \cmidrule(l){2-8}
        \multirow{2}*{Baseline}    & One-hot     & 0.69 & 0.68 & 0.72 & 0.52 & 0.53 & 0.58\\
                                   & Alignment & \bf{0.8} & \bf{0.76} & \bf{0.81} &\bf{ 0.63} & \bf{0.61} & \bf{0.68} \\
        \bottomrule
    \end{tabular}
\end{table}

We perform both three-class and eight-class secondary structure classification following the DSSP labeling system \cite{dssp}. Three way classification tags each position as either Helix, Strand or Other. Eight-way classification breaks these three labels into more specialized classes, for example Helix is broken into 3-turn, 4-turn or 5-turn helix. Table \ref{tab:secondary-structure} shows results on these tasks. We note that test-set performance is comparable for all three test sets, in particular alignment does better at both eight-way and three-way classification by a large margin. 

We follow the standard notation, where Q3 refers to three-way classification accuracy and Q8 refers to eight-way classification accuracy.

\subsubsection{Contact Prediction Results}
\label{sec:supplement-contact-results}

\begin{table}
  \caption{Detailed short-range contact prediction results. Short range contacts are contacts between positions separated by $6$ to $11$ positions, inclusive.}
  \label{tab:short-range-contact}
  \centering
  \begin{tabular}{llcccc}
    \toprule
    &  &  AUPRC  & P@L & P@L/2 & P@L/5\\
    \midrule
    \midrule
    \multirow{3}*{No Pretrain} & Transformer  &  0.29 & 0.25 & 0.32 & 0.4 \\
                               & LSTM         &  0.23 & 0.22 & 0.26 & 0.33 \\
                               & ResNet       &  0.2 & 0.18 & 0.24 & 0.31 \\
                               \cmidrule(l){2-6}
    \multirow{3}*{Pretrain}    & Transformer  &  0.35 & 0.28 & 0.35 & 0.46 \\
                               & LSTM         &  0.35 & 0.26 & 0.36 & 0.49\\
                               & ResNet      & 0.32 & 0.25 & 0.34 & 0.46\\
                               \cmidrule(l){2-6}
    Supervised \cite{bepler}   & LSTM         &  0.33 & 0.27 & 0.35 & 0.44 \\
    UniRep \cite{unirep}     & mLSTM          &  0.27 & 0.23 & 0.3 & 0.39 \\
                                \cmidrule(l){2-6}
    \multirow{2}*{Baseline}    & One-hot    &  0.3 & 0.26 & 0.34 & 0.42 \\
                               & Alignment  &  \bf{0.51} & \bf{0.35} & \bf{0.5} & \bf{0.66} \\
    \bottomrule
  \end{tabular}
\end{table}

\begin{table}
  \caption{Detailed medium-range contact prediction results. Medium range contacts are contacts between positions separated by $12$ to $23$ positions, inclusive.}
  \label{tab:medium-range-contact}
  \centering
  \begin{tabular}{llcccc}
    \toprule
    &   & AUPRC  & P@L & P@L/2 & P@L/5\\
    \midrule
    \midrule
    \multirow{3}*{No Pretrain} & Transformer & 0.2 & 0.18 & 0.24 & 0.31 \\
                               & LSTM        &  0.13 & 0.13 & 0.15 & 0.19 \\
                               & ResNet      & 0.15 & 0.14 & 0.18 & 0.23 \\
                               \cmidrule(l){2-6}
    \multirow{3}*{Pretrain}    & Transformer & 0.23 & 0.19 & 0.25 & 0.33\\
                               & LSTM        & 0.23 & 0.2 & 0.26 & 0.34\\
                               & ResNet      & 0.23 & 0.18 & 0.25 & 0.35\\
                               \cmidrule(l){2-6}
    Supervised \cite{bepler}   & LSTM        & 0.26 & 0.22 & 0.29 & 0.37 \\
    UniRep \cite{unirep}     & mLSTM         & 0.2 & 0.17 & 0.23 & 0.32 \\
                                \cmidrule(l){2-6}
    \multirow{2}*{Baseline}    & One-hot      & 0.2 & 0.17 & 0.23 & 0.3 \\
                               & Alignment & \bf{0.45} & \bf{0.32} & \bf{0.45} & \bf{0.59}\\
    \bottomrule
  \end{tabular}
\end{table}

\begin{table}
  \caption{Detailed long-range contact prediction results. Long range contacts are contacts between positions separated by $24$ or more positions, inclusive.}
  \label{tab:long-range-contact}
  \centering
  \begin{tabular}{llcccc}
    \toprule
    & & AUPRC  & P@L & P@L/2 & P@L/5\\
    \midrule
    \midrule
    \multirow{3}*{No Pretrain} & Transformer & 0.09 & 0.15 & 0.17 & 0.19 \\
                               & LSTM        & 0.05 & 0.1 & 0.12 & 0.15\\
                               & ResNet      & 0.06 & 0.11 & 0.13 & 0.15\\
                               \cmidrule(l){2-6}
    \multirow{3}*{Pretrain}    & Transformer & 0.1 & 0.17 & 0.2 & 0.24\\
                               & LSTM        & 0.11 & 0.2 & 0.23 & 0.27 \\
                               & ResNet      & 0.06 & 0.1 & 0.13 & 0.17 \\
                               \cmidrule(l){2-6}
    Supervised \cite{bepler}   & LSTM        & 0.11 & 0.18 & 0.22 & 0.26 \\
    UniRep \cite{unirep}     & mLSTM         & 0.09 & 0.17 & 0.2 & 0.22\\
                                \cmidrule(l){2-6}
    \multirow{2}*{Baseline}    & One-hot      & 0.07 & 0.13 & 0.16 & 0.22\\
                               & Alignment & \bf{0.2} & \bf{0.33} & \bf{0.42} & \bf{0.51} \\
    \bottomrule
  \end{tabular}
\end{table}

We report all metrics commonly used to capture contact prediction results \cite{raptorx} in tables \ref{tab:medium-range-contact} and \ref{tab:long-range-contact}. The metrics ``P@K'' are precision for the top $K$ contacts,  where all contacts are sorted from highest confidence to lowest confidence. Note that $L$ is the length of the protein, so ``P@L/2'', for example, denotes the precision for the $L/2$ most likely predicted contacts in a protein of length $L$. In Table \ref{tab:medium-range-contact} we report all metrics for medium range contacts, which are contacts for positions between five and twelve amino acids apart. In Table \ref{tab:long-range-contact} we report all metrics for long range contacts, which are contacts for positions greater than 12 amino acids apart. 

All results decay as we transition from short range to long range contacts, which we note is \textit{not} the case for many state of the art methods from recent CASP competitions \cite{raptorx, casp12-contact}.

\subsubsection{Remote Homology Results}
\label{sec:supplement-remote-homology-results}

\begin{table}
    \caption{Detailed remote homology prediction results}
    \label{tab:homology}
    \centering
    \begin{tabular}{llcccccc}
        \toprule
        \multicolumn{2}{c}{} & \multicolumn{2}{c}{Fold} & \multicolumn{2}{c}{Superfamily} & \multicolumn{2}{c}{Family}\\
        \cmidrule(l){3-4} \cmidrule(l){5-6} \cmidrule(l){7-8}
          & & Top-1 & Top-5 & Top-1 & Top-5 & Top-1 & Top-5\\
        \midrule
        \midrule
        \multirow{3}*{No Pretrain} & Transformer & 0.09 & 0.21 & 0.07 & 0.2 & 0.31 & 0.58 \\
                                   & LSTM        & 0.12 & 0.28 & 0.13 & 0.29 & 0.68 & 0.85 \\
                                   & ResNet      & 0.1 & 0.24 & 0.07 & 0.19 & 0.39 & 0.6 \\
                                   \cmidrule(l){2-8}
        \multirow{3}*{Pretrain}    & Transformer & 0.21 & 0.37 & 0.34 & 0.51 & 0.88 & 0.94 \\
                                   & LSTM        & \bf{0.26} & \bf{0.43} & \bf{0.43} & \bf{0.59} & \bf{0.92} & \bf{0.97} \\
                                   & ResNet      & 0.17 & 0.29 & 0.31 & 0.44 & 0.77 & 0.87 \\
                                   \cmidrule(l){2-8}
    Supervised \cite{bepler}   & LSTM        & 0.17 & 0.30 & 0.20 & 0.36 & 0.79 & 0.91 \\
    UniRep \cite{unirep}     & mLSTM         & 0.23 & 0.39 & 0.38 & 0.54 & 0.87 & 0.94 \\
                                    \cmidrule(l){2-8}
        \multirow{2}*{Baseline}    & One-hot      & 0.09 & 0.21 & 0.08 & 0.21 & 0.39 & 0.66 \\
                                   & Alignment & 0.09 & 0.21 & 0.09 & 0.24 & 0.53 & 0.77 \\
        \bottomrule
    \end{tabular}
\end{table}

In Table \ref{tab:homology}, we report results on three remote homology test datasets constructed in Hou et al \cite{deepsf}. Recall that ``Fold'' has the most distantly related proteins from train, while ``Superfamily'' and ``Family'' are increasingly related (see Section \ref{sec:supplement-remote-homology-details} for more details). This is reflected in the accuracies in Table \ref{tab:homology}, which increase drastically as the test sets get easier. 

\subsubsection{Fluorescence Results}
\label{sec:supplement-fluorescence-results} 

\begin{table}
    \caption{Detailed fluorescence prediction results. $\rho$ denotes Spearman $\rho$.}
    \label{tab:fluorescence}
    \centering
    \begin{tabular}{llcccccc}
        \toprule
        \multicolumn{2}{c}{} & \multicolumn{2}{c}{Full Test Set} & \multicolumn{2}{c}{Bright Mode Only} & \multicolumn{2}{c}{Dark Mode Only}\\
        \cmidrule(l){3-4} \cmidrule(l){5-6} \cmidrule(l){7-8}
          & & MSE &  $\rho$ & MSE & $\rho$ & MSE & $\rho$\\
        \midrule
        \midrule
        \multirow{3}*{No Pretrain} & Transformer & 2.59 & 0.22 & 0.08 & 0.08 & 3.79 & 0 \\
                                  & LSTM         & 2.35 & 0.21 & 0.11 & 0.05 & 3.43 & -0.01 \\
                                  & ResNet       & 2.79 & -0.28 & \bf{0.07} & -0.07 & 4.1 & -0.01 \\
                                  \cmidrule(l){2-8}
        \multirow{3}*{Pretrain}    & Transformer & 0.22 & \bf{0.68} & 0.09 & 0.60 & 0.29 & \bf{0.05} \\
                                  & LSTM         & \bf{0.19} & 0.67 & 0.12 & 0.62 & \bf{0.22} & 0.04 \\
                                  & ResNet       & 3.04 & 0.21 & 0.12 & 0.05 & 4.45 & 0.02 \\
                                  \cmidrule(l){2-8}
    Supervised \cite{bepler}   & LSTM        & 2.17 & 0.33 & 0.08 & 0.06 & 3.17 & 0.02 \\
    UniRep \cite{unirep}     & mLSTM         & 0.20 & 0.67 & 0.13 & \bf{0.63} & 0.24 & 0.04 \\
                                    \cmidrule(l){2-8}
        Baseline    & One-hot      & 2.69 & 0.14 & 0.08 & 0.03 & 3.95 & 0.0 \\
        \bottomrule
    \end{tabular}
\end{table}

Fluorescence distribution in the train, validation, and test sets is bimodal, with one mode corresponding to bright proteins and one mode corresponding to dark proteins. The dark mode is significantly more diverse in the test set than the train and validation sets, which makes sense as most random mutations will destroy the refined structure necessary for fluorescence. With this in  mind, we report Spearman's $\rho$ and mean-squared-error (MSE) on the whole test-set, on only dark mode, and on only the bright mode in Table \ref{tab:fluorescence}. The drop in MSE for both modes shows that pretraining helps our best models distinguish between dark and bright proteins. However low Spearman's $\rho$ for the dark mode suggests that models are not able to rank proteins within this mode.

\subsubsection{Stability Results} 
\label{sec:supplement-stability-results} 

\begin{table}
    \caption{Overall stability prediction results}
    \label{tab:stability-full-dataset}
    \centering
    \begin{tabular}{llcc}
        \toprule

         & & Spearman's $\rho$ & Accuracy \\
        \midrule
        \midrule
        \multirow{3}*{No Pretrain} & Transformer & -0.06&0.5\\
                                   & LSTM        & 0.28&0.6\\
                                   & ResNet      & 0.61&0.68\\
                                   \cmidrule(l){2-4}
        \multirow{3}*{Pretrain}    & Transformer & \bf{0.73}&\bf{0.70}\\
                                   & LSTM     
                                   & 0.69&0.69\\
                                   & ResNet      & \bf{0.73}&0.66\\
                                   \cmidrule(l){2-4}
    Supervised \cite{bepler}   & LSTM        &0.64&0.67\\
    UniRep \cite{unirep}     & mLSTM         & \bf{0.73}&0.69\\
                                    \cmidrule(l){2-4}
        Baseline                   & One-hot      & 0.19&0.58\\
        \bottomrule
    \end{tabular}
\end{table}

\begin{table}[H]
    \caption{Stability prediction results broken down by protein topology}
    \label{tab:stability-topologies}
    \centering
    \begin{tabular}{llcccccccc}
        \toprule
        \multicolumn{2}{c}{}  &
        \multicolumn{2}{c}{$\alpha \alpha \alpha$} & \multicolumn{2}{c}{$\alpha \beta \beta \alpha$} & 
        \multicolumn{2}{c}{$\beta \alpha \beta \beta$} & 
        \multicolumn{2}{c}{$\beta \beta \alpha \beta \beta$} 
        \\
        \cmidrule(l){3-4} \cmidrule(l){5-6} \cmidrule(l){7-8} \cmidrule(l){9-10}
         & & $\rho$ & Acc & $\rho$ & Acc & $\rho$ & Acc & $\rho$ & Acc \\
        \midrule
        \midrule
        \multirow{3}*{No Pretrain} & Transformer & -0.39&0.49&-0.41&0.47&0.52&0.5&0.25&0.52 \\
                                   & LSTM        &-0.07&0.57&0.39&0.7&-0.43&0.56&-0.34&0.56 \\
                                   & ResNet      & 0.64&0.69&0.16&0.69&0.63&0.67&0.65&0.67\\
                                   \cmidrule(l){2-10}
        \multirow{3}*{Pretrain}    & Transformer & 0.66&0.68&\bf{0.48}&0.73&0.65&\bf{0.71}&0.65&0.67\\
                                   & LSTM        & 0.71&\bf{0.7}&0.17&0.73&\bf{0.68}&0.67&\bf{0.67}&\bf{0.7} \\
                                   & ResNet      & 0.68&0.68&0.15&0.63&0.61&0.68&0.6&0.68\\
                                   \cmidrule(l){2-10}
    Supervised \cite{bepler}   & LSTM        & 0.33&0.66&0.24&\bf{0.79}&0.54&0.7&0.58&0.53 \\
    UniRep \cite{unirep}     & mLSTM         &\bf{0.72}&0.66&0.11&0.76&\bf{0.68}&0.66&0.65&0.67\\
                                    \cmidrule(l){2-10}
        Baseline                   & One-hot      & 0.58&0.59&0.04&0.58&-0.05&0.58&0.54&0.58\\
        \bottomrule
    \end{tabular}
\end{table}

The goal of the Rocklin \etal \cite{stability-paper} experiment was to find highly stable proteins. In the last stage of this experiment they examine variants of the the most promising candidate proteins. Therefore we wish to measure both whether our model was able to learn the landscape around these candidate proteins, as well as whether it successfully identified those variants with greater stability than the original parent proteins. In Table \ref{tab:stability-full-dataset} we report Spearman's $\rho$ to measure the degree to which the landscape was learned. In addition, we report classification accuracy of whether a mutation is beneficial or harmful using the predicted stability of the parent protein as a decision boundary.

In Table \ref{tab:stability-topologies} report all metrics separately for each of the four protein topologies tested in Rocklin et al \cite{stability-paper}, where $\alpha$ denotes a helix and $\beta$ denotes a strand (or $\beta$-sheet). We do this because success rates varied significantly by topology in their experiments, so some topologies (such as $\alpha \alpha \alpha$ were much easier to optimize than others (such as $\alpha \beta \beta \alpha$). We find that our prediction success also varies significantly by topology.

%% file: main.bbl
\begin{thebibliography}{10}

\bibitem{uniprot}
The~UniProt Consortium.
\newblock {UniProt: a worldwide hub of protein knowledge}.
\newblock {\em Nucleic Acids Research}, 47(D1):D506--D515, 2018.

\bibitem{anfinsen}
C~B Anfinsen, E~Haber, M~Sela, F~H White, and Jr.
\newblock {The kinetics of formation of native ribonuclease during oxidation of
  the reduced polypeptide chain.}
\newblock {\em Proceedings of the National Academy of Sciences of the United
  States of America}, 47(9):1309--14, 1961.

\bibitem{yanofsky}
C~Yanofsky, V~Horn, and D~Thorpe.
\newblock {Protein Structure Relationships Revealed by Mutational Analysis.}
\newblock {\em Science (New York, N.Y.)}, 146(3651):1593--4, 1964.

\bibitem{altschuh}
D~Altschuh, T~Vernet, P~Berti, D~Moras, and K~Nagai.
\newblock {Coordinated amino acid changes in homologous protein families.}
\newblock {\em Protein engineering}, 2(3):193--9, 1988.

\bibitem{elmo}
Matthew Peters, Mark Neumann, Mohit Iyyer, Matt Gardner, Christopher Clark,
  Kenton Lee, and Luke Zettlemoyer.
\newblock {Deep Contextualized Word Representations}.
\newblock In {\em Proceedings of the 2018 Conference of the North
  {\{}A{\}}merican Chapter of the Association for Computational Linguistics:
  Human Language Technologies, Volume 1 (Long Papers)}, pages 2227--2237, New
  Orleans, Louisiana, 2018. Association for Computational Linguistics.

\bibitem{bert}
Jacob Devlin, Ming-Wei Chang, Kenton Lee, and Kristina Toutanova.
\newblock {BERT: Pre-training of Deep Bidirectional Transformers for Language
  Understanding}.
\newblock {\em arXiv}, 2018.

\bibitem{gpt2}
Alec Radford, Jeffrey Wu, Rewon Child, David Luan, Dario Amodei, and Ilya
  Sutskever.
\newblock Language models are unsupervised multitask learners.
\newblock {\em OpenAI Blog}, 1:8, 2019.

\bibitem{antibodyengineering}
Timothy~A Whitehead, Aaron Chevalier, Yifan Song, Cyrille Dreyfus, Sarel~J
  Fleishman, Cecilia De~Mattos, Chris~A Myers, Hetunandan Kamisetty, Patrick
  Blair, Ian~A Wilson, et~al.
\newblock Optimization of affinity, specificity and function of designed
  influenza inhibitors using deep sequencing.
\newblock {\em Nature biotechnology}, 30(6):543, 2012.

\bibitem{pancreatic-cancer-paper}
Esther Vazquez, Neus Ferrer-Miralles, Ramon Mangues, Jose~L Corchero,
  Jr~Schwartz, Antonio Villaverde, et~al.
\newblock Modular protein engineering in emerging cancer therapies.
\newblock {\em Current pharmaceutical design}, 15(8):893--916, 2009.

\bibitem{darkproteome}
Nelson Perdig{\~a}o, Julian Heinrich, Christian Stolte, Kenneth~S. Sabir,
  Michael~J. Buckley, Bruce Tabor, Beth Signal, Brian~S. Gloss, Christopher~J.
  Hammang, Burkhard Rost, Andrea Schafferhans, and Se{\'a}n~I.
  O{\textquoteright}Donoghue.
\newblock Unexpected features of the dark proteome.
\newblock {\em Proceedings of the National Academy of Sciences},
  112(52):15898--15903, 2015.

\bibitem{bepler}
Tristan Bepler and Bonnie Berger.
\newblock Learning protein sequence embeddings using information from
  structure.
\newblock In {\em International Conference on Learning Representations}, 2019.

\bibitem{unirep}
Ethan~C. Alley, Grigory Khimulya, Surojit Biswas, Mohammed AlQuraishi, and
  George~M. Church.
\newblock {Unified rational protein engineering with sequence-only deep
  representation learning}.
\newblock {\em bioRxiv}, page 589333, 2019.

\bibitem{psi-blast}
Stephen~F Altschul, Thomas~L Madden, Alejandro~A Sch{\"a}ffer, Jinghui Zhang,
  Zheng Zhang, Webb Miller, and David~J Lipman.
\newblock Gapped blast and psi-blast: a new generation of protein database
  search programs.
\newblock {\em Nucleic acids research}, 25(17):3389--3402, 1997.

\bibitem{hhblits}
Michael Remmert, Andreas Biegert, Andreas Hauser, and Johannes S{\"{o}}ding.
\newblock {HHblits: lightning-fast iterative protein sequence searching by
  HMM-HMM alignment}.
\newblock {\em Nature Methods}, 9(2):173--175, 2012.

\bibitem{hhpred}
J.~Soding, A.~Biegert, and A.~N. Lupas.
\newblock {The HHpred interactive server for protein homology detection and
  structure prediction}.
\newblock {\em Nucleic Acids Research}, 33(Web Server):W244--W248, 2005.

\bibitem{profilehmm}
Sean~R. Eddy.
\newblock Profile hidden markov models.
\newblock {\em Bioinformatics (Oxford, England)}, 14(9):755--763, 1998.

\bibitem{IUPACnomenclature}
IUPAC-IUB.
\newblock Recommendations on nomenclature and symbolism for amino acids and
  peptides.
\newblock {\em Pure Appl. Chem}, 56:595--623, 1984.

\bibitem{creightonproteins}
Thomas~E Creighton.
\newblock {\em Proteins: structures and molecular properties}.
\newblock Macmillan, 1993.

\bibitem{rost-sequence-alignment}
Burkhard Rost.
\newblock {Twilight zone of protein sequence alignments}.
\newblock {\em Protein Engineering, Design and Selection}, 12(2):85--94, 1999.

\bibitem{brenner1998}
Steven~E Brenner, Cyrus Chothia, and Tim~JP Hubbard.
\newblock Assessing sequence comparison methods with reliable structurally
  identified distant evolutionary relationships.
\newblock {\em Proceedings of the National Academy of Sciences},
  95(11):6073--6078, 1998.

\bibitem{blast}
Stephen~F. Altschul, Warren Gish, Webb Miller, Eugene~W. Myers, and David~J.
  Lipman.
\newblock {Basic local alignment search tool}.
\newblock {\em Journal of Molecular Biology}, 215(3):403--410, 1990.

\bibitem{durbin}
Richard Durbin, Sean~R Eddy, Anders Krogh, and Graeme Mitchison.
\newblock {\em Biological sequence analysis: probabilistic models of proteins
  and nucleic acids}.
\newblock Cambridge university press, 1998.

\bibitem{smith-waterman}
Temple~F Smith, Michael~S Waterman, et~al.
\newblock Identification of common molecular subsequences.
\newblock {\em Journal of molecular biology}, 147(1):195--197, 1981.

\bibitem{arnold2006swiss}
Konstantin Arnold, Lorenza Bordoli, J{\"u}rgen Kopp, and Torsten Schwede.
\newblock The swiss-model workspace: a web-based environment for protein
  structure homology modelling.
\newblock {\em Bioinformatics}, 22(2):195--201, 2006.

\bibitem{chapelle2009semi}
Olivier Chapelle, Bernhard Scholkopf, and Alexander Zien.
\newblock Semi-supervised learning.
\newblock {\em IEEE Transactions on Neural Networks}, 20(3):542--542, 2009.

\bibitem{casp}
John Moult, Krzysztof Fidelis, Andriy Kryshtafovych, Torsten Schwede, and Anna
  Tramontano.
\newblock {Critical assessment of methods of protein structure prediction
  (CASP)-Round XII}.
\newblock {\em Proteins: Structure, Function, and Bioinformatics}, 86:7--15,
  2018.

\bibitem{proteinnet}
Mohammed AlQuraishi.
\newblock {ProteinNet: a standardized data set for machine learning of protein
  structure}.
\newblock {\em bioRxiv}, 2019.

\bibitem{protein-cluster-kernels}
Jason Weston, Dengyong Zhou, Andr{\'e} Elisseeff, William~S Noble, and
  Christina~S Leslie.
\newblock Semi-supervised protein classification using cluster kernels.
\newblock In {\em Advances in neural information processing systems}, pages
  595--602, 2004.

\bibitem{shin2006prediction}
Hyunjung Shin, Koji Tsuda, B~Sch{\"o}lkopf, A~Zien, et~al.
\newblock Prediction of protein function from networks.
\newblock In {\em Semi-supervised learning}, pages 361--376. MIT press, 2006.

\bibitem{seqvec}
Michael Heinzinger, Ahmed Elnaggar, Yu~Wang, Christian~4 Dallago, Dmitrii
  Nachaev, Florian Matthes, and {\&}~Burkhard Rost.
\newblock {Modeling the Language of Life - Deep Learning Protein Sequences}.
\newblock {\em bioRxiv}, 2019.

\bibitem{fair-bert}
Alexander Rives, Siddharth Goyal, Joshua Meier, Demi Guo, Myle Ott, C.~Lawrence
  Zitnick, Jerry Ma, and Rob Fergus.
\newblock Biological structure and function emerge from scaling unsupervised
  learning to 250 million protein sequences.
\newblock {\em bioRxiv}, 2019.

\bibitem{deepsequence}
Adam~J. Riesselman, John~B. Ingraham, and Debora~S. Marks.
\newblock {Deep generative models of genetic variation capture the effects of
  mutations}.
\newblock {\em Nature Methods}, 15(10):816--822, 2018.

\bibitem{arnold-embedding}
Kevin~K Yang, Zachary Wu, Claire~N Bedbrook, and Frances~H Arnold.
\newblock Learned protein embeddings for machine learning.
\newblock {\em Bioinformatics}, 34(15):2642--2648, 2018.

\bibitem{superglue}
Alex Wang, Yada Pruksachatkun, Nikita Nangia, Amanpreet Singh, Julian Michael,
  Felix Hill, Omer Levy, and Samuel~R Bowman.
\newblock Superglue: A stickier benchmark for general-purpose language
  understanding systems.
\newblock {\em arXiv preprint arXiv:1905.00537}, 2019.

\bibitem{pfam}
Sara El-Gebali, Jaina Mistry, Alex Bateman, Sean~R Eddy, Aur{\'{e}}lien
  Luciani, Simon~C Potter, Matloob Qureshi, Lorna~J Richardson, Gustavo~A
  Salazar, Alfredo Smart, Erik~L L Sonnhammer, Layla Hirsh, Lisanna Paladin,
  Damiano Piovesan, Silvio~C E Tosatto, and Robert~D Finn.
\newblock {The Pfam protein families database in 2019}.
\newblock {\em Nucleic Acids Research}, 47(D1):D427--D432, 2019.

\bibitem{netsurfp}
Michael~Schantz Klausen, Martin~Closter Jespersen, Henrik Nielsen,
  Kamilla~Kjaergaard Jensen, Vanessa~Isabell Jurtz, Casper~Kaae Soenderby,
  Morten Otto~Alexander Sommer, Ole Winther, Morten Nielsen, Bent Petersen,
  et~al.
\newblock Netsurfp-2.0: Improved prediction of protein structural features by
  integrated deep learning.
\newblock {\em Proteins: Structure, Function, and Bioinformatics}, 2019.

\bibitem{jpred}
Alexey Drozdetskiy, Christian Cole, James Procter, and Geoffrey~J. Barton.
\newblock {JPred4: a protein secondary structure prediction server}.
\newblock {\em Nucleic Acids Research}, 43(W1):W389--W394, 2015.

\bibitem{cb513}
James~A Cuff and Geoffrey~J Barton.
\newblock Evaluation and improvement of multiple sequence methods for protein
  secondary structure prediction.
\newblock {\em Proteins: Structure, Function, and Bioinformatics},
  34(4):508--519, 1999.

\bibitem{kim}
David~E. Kim, Frank DiMaio, Ray {Yu-Ruei Wang}, Yifan Song, and David Baker.
\newblock {One contact for every twelve residues allows robust and accurate
  topology-level protein structure modeling}.
\newblock {\em Proteins: Structure, Function, and Bioinformatics}, 82:208--218,
  2014.

\bibitem{deepsf}
Jie Hou, Badri Adhikari, and Jianlin Cheng.
\newblock Deepsf: deep convolutional neural network for mapping protein
  sequences to folds.
\newblock {\em Bioinformatics}, 34(8):1295--1303, 2017.

\bibitem{amr-paper}
Leticia~Stephan Tavares, Carolina dos Santos Fernandes~da Silva,
  Vinicius~Carius Souza, V{\^a}nia L{\'u}cia~da Silva, Cl{\'a}udio~Galuppo
  Diniz, and Marcelo De~Oliveira Santos.
\newblock Strategies and molecular tools to fight antimicrobial resistance:
  resistome, transcriptome, and antimicrobial peptides.
\newblock {\em Frontiers in microbiology}, 4:412, 2013.

\bibitem{casx}
Jun-Jie Liu, Natalia Orlova, Benjamin~L Oakes, Enbo Ma, Hannah~B Spinner,
  Katherine~LM Baney, Jonathan Chuck, Dan Tan, Gavin~J Knott, Lucas~B
  Harrington, et~al.
\newblock Casx enzymes comprise a distinct family of rna-guided genome editors.
\newblock {\em Nature}, 566(7743):218, 2019.

\bibitem{gfp-paper}
Karen~S Sarkisyan, Dmitry~A Bolotin, Margarita~V Meer, Dinara~R Usmanova,
  Alexander~S Mishin, George~V Sharonov, Dmitry~N Ivankov, Nina~G Bozhanova,
  Mikhail~S Baranov, Onuralp Soylemez, et~al.
\newblock Local fitness landscape of the green fluorescent protein.
\newblock {\em Nature}, 533(7603):397, 2016.

\bibitem{arnold-ml-review}
Kevin~K Yang, Zachary Wu, and Frances~H Arnold.
\newblock Machine learning in protein engineering.
\newblock {\em arXiv preprint arXiv:1811.10775}, 2018.

\bibitem{stability-paper}
Gabriel~J Rocklin, Tamuka~M Chidyausiku, Inna Goreshnik, Alex Ford, Scott
  Houliston, Alexander Lemak, Lauren Carter, Rashmi Ravichandran, Vikram~K
  Mulligan, Aaron Chevalier, et~al.
\newblock Global analysis of protein folding using massively parallel design,
  synthesis, and testing.
\newblock {\em Science}, 357(6347):168--175, 2017.

\bibitem{mikolov2010recurrent}
Tom{\'a}{\v{s}} Mikolov, Martin Karafi{\'a}t, Luk{\'a}{\v{s}} Burget, Jan
  {\v{C}}ernock{\`y}, and Sanjeev Khudanpur.
\newblock Recurrent neural network based language model.
\newblock In {\em Eleventh annual conference of the international speech
  communication association}, 2010.

\bibitem{decaf}
Jeff Donahue, Yangqing Jia, Oriol Vinyals, Judy Hoffman, Ning Zhang, Eric
  Tzeng, and Trevor Darrell.
\newblock {DeCAF: A Deep Convolutional Activation Feature for Generic Visual
  Recognition}.
\newblock {\em Journal of Machine Learning Research}, 2013.

\bibitem{lstm}
Sepp Hochreiter and J{\"u}rgen Schmidhuber.
\newblock Long short-term memory.
\newblock {\em Neural computation}, 9(8):1735--1780, 1997.

\bibitem{vaswani}
Ashish Vaswani, Noam Shazeer, Niki Parmar, Jakob Uszkoreit, Llion Jones,
  Aidan~N Gomez, {\L}ukasz Kaiser, and Illia Polosukhin.
\newblock {Attention is All you Need}.
\newblock In I~Guyon, U~V Luxburg, S~Bengio, H~Wallach, R~Fergus,
  S~Vishwanathan, and R~Garnett, editors, {\em Advances in Neural Information
  Processing Systems 30}, pages 5998--6008. Curran Associates, Inc., 2017.

\bibitem{dilated-residual-networks}
Fisher Yu, Vladlen Koltun, and Thomas Funkhouser.
\newblock Dilated residual networks.
\newblock In {\em The IEEE Conference on Computer Vision and Pattern
  Recognition (CVPR)}, 2017.

\bibitem{mlstm}
Ben Krause, Liang Lu, Iain Murray, and Steve Renals.
\newblock Multiplicative lstm for sequence modelling.
\newblock {\em arXiv preprint arXiv:1609.07959}, 2016.

\bibitem{casp12-contact}
Joerg Schaarschmidt, Bohdan Monastyrskyy, Andriy Kryshtafovych, and
  Alexandre~MJJ Bonvin.
\newblock Assessment of contact predictions in casp12: Co-evolution and deep
  learning coming of age.
\newblock {\em Proteins: Structure, Function, and Bioinformatics}, 86:51--66,
  2018.

\bibitem{raptorx}
Jianzhu Ma, Sheng Wang, Zhiyong Wang, and Jinbo Xu.
\newblock Protein contact prediction by integrating joint evolutionary coupling
  analysis and supervised learning.
\newblock {\em Bioinformatics}, 31(21):3506--3513, 2015.

\bibitem{casp12}
John Moult, Krzysztof Fidelis, Andriy Kryshtafovych, Torsten Schwede, and Anna
  Tramontano.
\newblock Critical assessment of methods of protein structure prediction
  (casp)—round xii.
\newblock {\em Proteins: Structure, Function, and Bioinformatics}, 86:7--15,
  2018.

\bibitem{ts115}
Yuedong Yang, Jianzhao Gao, Jihua Wang, Rhys Heffernan, Jack Hanson, Kuldip
  Paliwal, and Yaoqi Zhou.
\newblock Sixty-five years of the long march in protein secondary structure
  prediction: the final stretch?
\newblock {\em Briefings in bioinformatics}, 19(3):482--494, 2016.

\bibitem{structure-cost}
Raymond~C Stevens.
\newblock The cost and value of three-dimensional protein structure.
\newblock {\em Drug Discovery World}, 4(3):35--48, 2003.

\bibitem{scope}
Naomi~K Fox, Steven~E Brenner, and John-Marc Chandonia.
\newblock Scope: Structural classification of proteins—extended, integrating
  scop and astral data and classification of new structures.
\newblock {\em Nucleic acids research}, 42(D1):D304--D309, 2013.

\bibitem{pdb}
Helen~M Berman, John Westbrook, Zukang Feng, Gary Gilliland, Talapady~N Bhat,
  Helge Weissig, Ilya~N Shindyalov, and Philip~E Bourne.
\newblock The protein data bank.
\newblock {\em Nucleic acids research}, 28(1):235--242, 2000.

\bibitem{new-bacterial-domains}
Cindy~J Castelle and Jillian~F Banfield.
\newblock Major new microbial groups expand diversity and alter our
  understanding of the tree of life.
\newblock {\em Cell}, 172(6):1181--1197, 2018.

\bibitem{ccmpred}
Stefan Seemayer, Markus Gruber, and Johannes S{\"{o}}ding.
\newblock {CCMpred—fast and precise prediction of protein residue–residue
  contacts from correlated mutations}.
\newblock {\em Bioinformatics}, 30(21):3128--3130, 2014.

\bibitem{memory-saving-gradients}
Tianqi Chen, Bing Xu, Chiyuan Zhang, and Carlos Guestrin.
\newblock {Training Deep Nets with Sublinear Memory Cost}.
\newblock {\em arXiv}, 2016.

\bibitem{dssp}
Robbie~P Joosten, Tim~AH Te~Beek, Elmar Krieger, Maarten~L Hekkelman, Rob~WW
  Hooft, Reinhard Schneider, Chris Sander, and Gert Vriend.
\newblock A series of pdb related databases for everyday needs.
\newblock {\em Nucleic acids research}, 39(suppl\_1):D411--D419, 2010.

\end{thebibliography}
